\documentclass[lettersize,journal]{IEEEtran}
\usepackage{times}
\usepackage{comment}

\usepackage[numbers]{natbib}
\usepackage{booktabs, multirow, multicol}
\usepackage{graphicx}
\usepackage{subcaption}
\usepackage{amsmath,amsthm,amsfonts,amssymb,amscd}
\usepackage[ruled,vlined,noend,linesnumbered]{algorithm2e}
\usepackage[bookmarks=true]{hyperref}
\usepackage{makecell}

\usepackage[normalem]{ulem}
\usepackage[disable]{todonotes}

\newcommand{\wenzhen}[1]{\todo[inline,color=red!40]{Wenzhen: #1}}
\newcommand{\joe}[1]{\todo[inline,color=blue!20]{Joe: #1}}
\newcommand{\amin}[1]{\todo[inline,color=purple!20]{Amin: #1}}

\newcommand{\revised}[1]{\textcolor{black}{#1}}
\newcommand{\mytodo}[1]{\textcolor{red}{TODO: #1}}

\DeclareMathOperator*{\argmin}{arg\,min}

\title{GelSLAM: A Real-time, High-Fidelity, and Robust \\ 3D Tactile SLAM System}

\author{Hung-Jui Huang$^{1}$, Mohammad Amin Mirzaee$^{2}$, Michael Kaess$^{1}$, and Wenzhen Yuan$^{2}$% <-this % stops a space
\thanks{$^{1}$Hung-Jui Huang and Michael Kaess are with Carnegie Mellon University, Pittsburgh, PA, USA 
        {\tt\small \{hungjuih, kaess\}@andrew.cmu.edu }}%
\thanks{$^{2}$Mohammad Amin Mirzaee and Wenzhen Yuan are with University of Illinois Urbana-Champaign, Champaign, IL, USA
        {\tt\small {mirzaee2, yuanwz}@illinois.edu }}%
}
\begin{document}



\maketitle
\thispagestyle{empty}
\pagestyle{empty}

\begin{abstract}
Accurately perceiving an object’s pose and shape is essential for precise grasping and manipulation. Compared to common vision-based methods, tactile sensing offers advantages in precision and immunity to occlusion when tracking and reconstructing objects in contact. This makes it particularly valuable for in-hand and other high-precision manipulation tasks. In this work, we present GelSLAM, a real-time 3D SLAM system that relies solely on tactile sensing to estimate object pose over long periods and reconstruct object shapes with high fidelity. Unlike traditional point cloud–based approaches, GelSLAM uses tactile-derived surface normals and curvatures for robust tracking and loop closure. It can track object motion in real time with low error and minimal drift, and reconstruct shapes with submillimeter accuracy, even for low-texture objects such as wooden tools. GelSLAM extends tactile sensing beyond local contact to enable global, long-horizon spatial perception, and we believe it will serve as a foundation for many precise manipulation tasks involving interaction with objects in hand. The video demo, code, and dataset are available at https://joehjhuang.github.io/gelslam.
\end{abstract}

\begin{comment}
\begin{IEEEkeywords}
Force and tactile sensing, 
perception for grasping and manipulation.
\end{IEEEkeywords}
\end{comment}

\section{Introduction}
\begin{comment}
In the history of robotics, breakthroughs in perception have often enabled breakthroughs in capability. Robust localization \cite{dellaert99} and environment mapping \cite{zhang14} with LiDAR have powered the development of autonomous driving. Dense 3D reconstruction \cite{li23} \cite{newcombe11} \cite{wang21} and robot pose estimation \cite{murartal15} \cite{schoenberger16} with vision have transformed indoor navigation and AR/VR. But for tactile perception, such a turning point has not yet arrived. Despite the rise of high-resolution tactile sensors \cite{lambeta24} \cite{yuan17}, tactile sensing has remained largely in a supporting role for object-centric perception (object pose estimation and 3D reconstruction) \cite{li25} \cite{suresh24} \cite{zhao23}. While object-centric perception is important for robot manipulation and other applications, including AR/VR and geometric study of objects in fields like biology, geology, and archaeology, tactile sensing is often used only to compensate for visual occlusion or refine vision-based estimates. Standalone tactile systems \cite{sodhi22} \cite{suresh21} \cite{wang18} \cite{zhao23} are generally known to perform significantly worse than their visual counterparts.
\end{comment}
%, one of the five human senses,
%\IEEEPARstart{T}{actile}
Tactile sensing plays a critical role in how humans perceive and interact with the world \cite{lederman87}. Through touch, we can effortlessly infer object properties ranging from physical ones such as weight and hardness to geometric ones like shape and pose \cite{lederman93}. In particular, geometric perception through touch enables object-level spatial understanding. For example, humans can reach into a bag and recognize the shapes of objects by feel, navigate in the dark by touching their surroundings, or sense the pose of a pen in hand while writing. However, enabling robots to achieve comparable spatial understanding through touch remains a significant challenge.
\joe{Dan: citing survey papers here?}
\wenzhen{Your work is only about geometric-based tracking/reconstruction, so you don't need to emphasize on the property-based perception}

\begin{comment}
\wenzhen{I think this definition is vague. You either want to use some more specific phrase or add another sentence to explain more} \wenzhen{somewhere you should motivate why "tactile only"} \wenzhen{I think you need more explaination of what's the raw format of tactile signals, then how is that connected to point cloud. Particularly, since you didn't explicitly define "tactile sensing", they could mean the form of traditional array-based sensors, or things like GelSight, or things like TacTip. Somewhere you want to make it explicit whether you are addressing tactile signals in general or some particular type of tactile signals} \wenzhen{without clear deinitation, this explanation might be hard to understand by people who are not tactile sensing experts} \wenzhen{I'm confused here because GelSight also has this limitation that it cannot capture the deep geometries. So your method didn't solve this problem}
\wenzhen{I think the motivation from the deficiency of current literature is weak.}
\end{comment}

\joe{I think mentioning the scientific motivation is also important, which is extending touch from a local sensing modality to one capable of global, long-horizon spatial understanding.}
%Our long-term goal is to understand how tactile sensing can enable robust and accurate object-level spatial understanding. \wenzhen{It's only one paper so your long-term goal is not that important here, except that you are talking about something directly relevant }
In this work, we aim to achieve real-time long-horizon object tracking and high-fidelity object 3D reconstruction using tactile input alone. 
\revised{Real-time object tracking using touch is particularly useful for contact-rich manipulation with robot hands, where high precision is required but vision is often occluded or suffers from low lighting. Meanwhile, object 3D reconstruction using tactile sensing benefits applications in biology, archaeology, dentistry, and geology that require capturing fine surface details often missed by vision.} We focus on GelSight-related tactile sensors \cite{lambeta24, yuan17}, which are known for their high spatial resolution for capturing local surface geometry. 

\begin{figure}[htbp]
\centering
\includegraphics[width=0.95\linewidth]{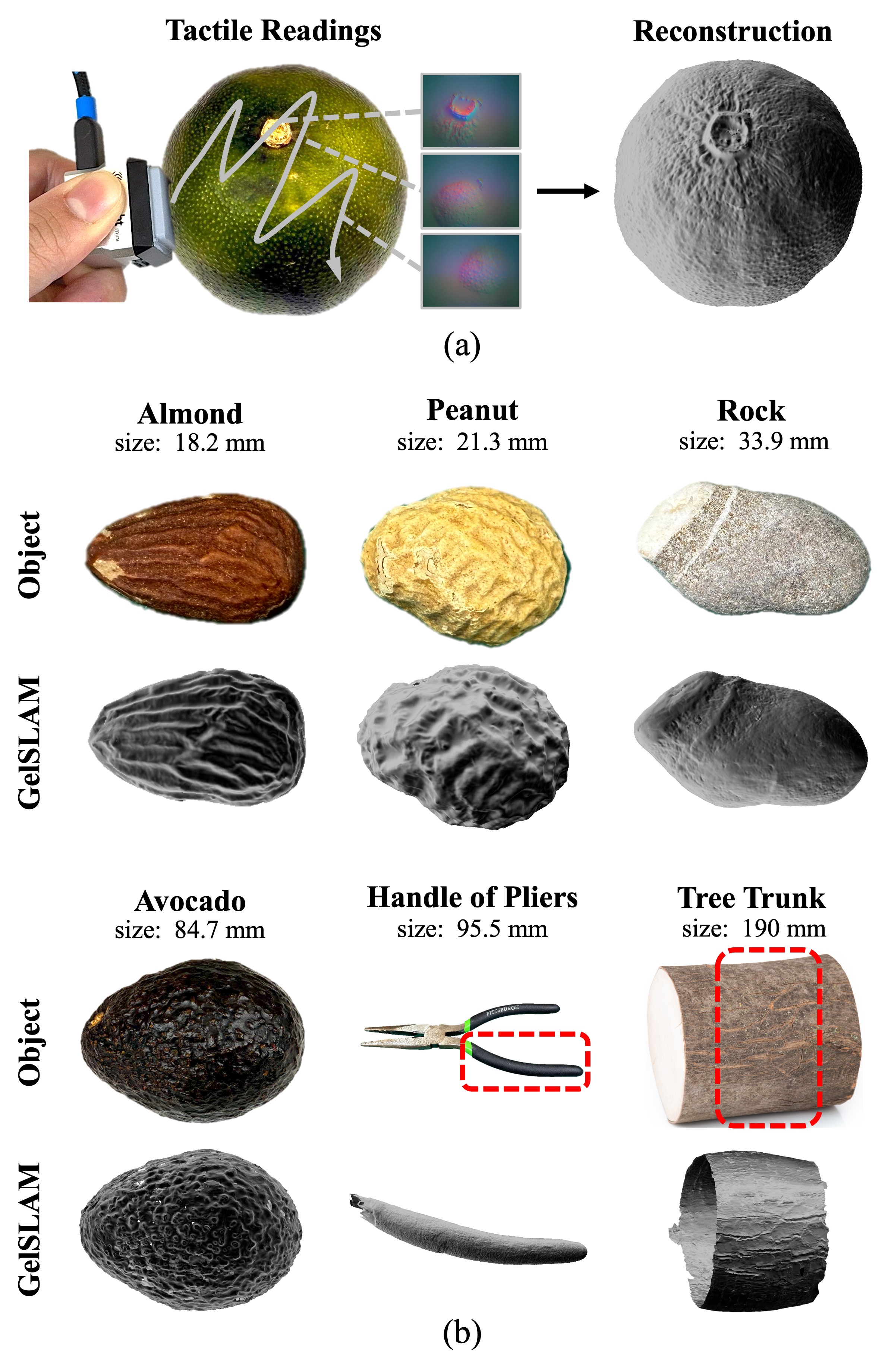}
\caption{GelSLAM enables robust, high-fidelity object-level 3D reconstruction and real-time, accurate long-horizon object tracking using only tactile sensing, as shown in (a). (b) Reconstruction results on a wide variety of objects, including small items like almonds and peanuts, low-texture objects such as the handle of pliers, and large objects like a tree trunk.\wenzhen{I think (b) is redundant}}
\label{fig:teaser}
\end{figure}

\begin{comment}
 We aim to achieve real-time, high-fidelity, and reliable object-centric perception on common objects using only tactile sensing. Previous work typically treats tactile data as point clouds and applies methods developed for point cloud data, including geometric registration (e.g., ICP for tracking \cite{sodhi22} \cite{wang21_1}), feature extraction (e.g., FPFH for pose estimation \cite{lu23}), and reconstruction \cite{suresh21} \cite{suresh24}. However, tactile point clouds suffer from a fundamental limitation: unlike vision or LiDAR, tactile depth is confined to the contact surface, resulting in point clouds that are inherently shallow and nearly planar. This makes tactile point clouds poorly suited for object-centric perception tasks. For example, ICP-based tracking methods often lose track due to insufficient surface variation, and FPFH features are typically unreliable unless applied to 3D-printed objects with exaggerated depth. 
\end{comment}

The key challenge is the ``blind men and the elephant" problem: each tactile reading provides rich but local information, and forming spatial relationships between them to construct a coherent global understanding is difficult without external cues. As a result, standalone tactile systems \cite{sodhi22, suresh2022shapemap, wang18, zhao23} are generally known to perform significantly worse than their visual counterparts \cite{li23, newcombe11, wang21, bowen24} in tracking and reconstruction tasks. To infer the relative motion between contact frames, previous works typically represent GelSight images as point clouds and apply standard techniques from point cloud processing, including geometric registration (e.g., ICP  \cite{sodhi22, wang21_1}), feature extraction (e.g., FPFH \cite{lu23}), and reconstruction \cite{suresh2022shapemap, suresh24}. However, tactile contact produces only small surface deformations, typically just a few millimeters. This leads to point clouds that are mostly flat and often lack distinctive geometric features, making them poorly suited for reliable alignment. For instance, ICP-based tracking methods frequently lose track due to insufficient surface variation, and FPFH features are typically unreliable unless applied to 3D-printed objects with exaggerated texture depth.

Our key insight for estimating spatial transformations is to treat GelSight images not as point clouds, but as \textbf{differential representations}: normal maps (first-order surface geometry) \cite{huang24} and curvature maps (second-order). While tactile point clouds have minimal depth variation, their differential representations can be rich in local features. For example, textured fabric may produce a nearly planar point cloud, yet its normal map captures fine surface textures such as weaves or ridges. Another advantage is that GelSight sensors directly measure normal maps via photometric stereo \cite{wang21_1}, while interpreting data as point clouds requires depth reconstruction through integration, introducing accumulated noise and distortion.

Building on this insight, we introduce \textbf{GelSLAM}, a system for real-time, accurate long-horizon object tracking and high-fidelity object-level 3D reconstruction using only touch. By operating directly on differential representations, GelSLAM robustly estimates the object's pose over time and fuses local tactile geometry into a globally consistent 3D model. GelSLAM consists of three main modules. The tracking module estimates the object’s motion between close-by frames and selects keyframes. However, simply composing these short-range transformations to achieve long-horizon tracking leads to drift. To correct this, the loop closure module detects when the sensor revisits previously touched regions, even after long intervals or contact interruptions, and estimates the relative pose between those frames. Globally consistent object poses for all previous frames are then computed through pose graph optimization that integrates frame-to-frame tracking and loop closure results. The reconstruction module fuses local surface patches from each tactile reading at their estimated locations to build a global 3D model. All core components of GelSLAM rely on differential representations of surface geometry, including relative pose estimation in tracking and loop closure, as well as keyframe selection and outlier rejection.

\begin{comment}
GelSLAM consists of three main components: tracking, loop closure, and pose graph optimization. Leveraging differential representations, these modules operate far more robustly than point cloud-based methods.\wenzhen{This sentence doesn't flow well here} For tracking, we apply NormalFlow on normal maps, with keyframes selected based on curvature map discrepancies. Loop closure initializes NormalFlow using SIFT feature \cite{lowe04} matches on curvature maps and uses the same curvature-based discrepancy check to reject outliers. Pose graph optimization fuses results from tracking and loop closures to compute globally consistent object poses online while minimizing drift. Finally, GelSLAM reconstructs the object by pasting local tactile 3D patches at their estimated poses, producing a high-fidelity model.\wenzhen{I think you need simpler definition of each modules for people who are not familiar with SLAM} 
\end{comment}

We evaluate GelSLAM on long-horizon object tracking and object-level 3D reconstruction across many common objects. In tracking, GelSLAM consistently outperforms prior methods, including NormalFlow \cite{huang24}, point cloud registration \cite{chen91, wei19}, and full SLAM pipelines such as Tac2Structure \cite{lu23}. We provide a detailed analysis of how each component of GelSLAM contributes to these improvements. For 3D reconstruction, GelSLAM achieves high-fidelity results even on objects with small surface texture, such as wood-made items, accurately estimating frame poses and producing remarkably detailed 3D models. It reconstructs objects with submillimeter-level Chamfer Distance compared to ground-truth geometry. Additionally, with the GelBelt sensor \cite{mirzaee25}, GelSLAM can reconstruct large objects such as a tree trunk while preserving fine surface details.

\revised{\bf{Contributions:}} \revised{Our main contribution is the three modules, each incorporating novel components tailored to tactile sensing, as well as the well-engineered system that integrates them into a robust tactile SLAM framework. The novel tactile-oriented components include tracking failure detection, keyframe selection, and robust loop detection. Together, these components enable GelSLAM to operate reliably in real time over tens of thousands of frames, whereas prior methods scale only to hundreds of frames and fail frequently.} To the best of our knowledge, GelSLAM is the first tactile-only system to achieve object-level 3D reconstruction with this degree of robustness, accuracy, and fidelity. GelSLAM shows that what was once a local sensing modality, touch, can move beyond its inherent limitations to support global, long-horizon spatial understanding. We believe it lays a foundation for future research in tactile-based perception, visuo-tactile integration, contact-rich manipulation, and dexterous manipulation. Beyond robotics, potential applications include dental scanning, biological phenotype analysis, in-the-wild geological surface inspection, archaeological reconstruction, and detailed shape capture for AR/VR. To support continued progress in the field, we release the full source code and dataset to the community. \joe{Remember to put URL here.}

\section{Related Work}
\wenzhen{somewhere here you should highlight the difference with NormalFlow}

\subsection{Object Pose Estimation}
Perceiving an object’s 6DoF pose is essential for many robotics tasks, including manipulation. The most common approach is through vision. When the object’s shape is known in advance, three main types of approaches are used. Keypoint-based methods \cite{peng19, sun22} detect 2D keypoints in an image and estimate pose from 2D–3D correspondences using a Perspective-n-Point (PnP) solver. Direct regression methods \cite{kehl17, xiang18} use neural networks to predict the pose directly from the input image. Iterative approaches \cite{iwase21, li18, wang19} refine an initial pose by repeatedly comparing rendered and observed images. When the object’s shape is unknown, the problem is often framed as object tracking, where recent works \cite{lin24, bowen24} reconstruct object representations from one or a few observations and track the pose over time.

Despite its popularity in pose estimation, vision often struggles in real-world robotics due to occlusions \cite{suresh24} and poor performance on optically challenging objects \cite{comi2024snapit}. Tactile sensing, on the other hand, offers high precision and is immune to occlusions as well as transparent or specular surfaces. When the object’s geometry is known, \cite{bimbo16} estimates a grasped object’s 6DoF pose by using a PCA-based covariance matching method to align the statistical structure of tactile array data with the local object geometry. Using a vision-based tactile sensor, DIGIT \cite{lambeta20}, Midas-Touch \cite{suresh22midas} applies a particle filter to track the evolving 6DoF pose distribution of a fingertip sliding across a known object’s surface. Tac2Pose \cite{bauza23} estimates the pose of a known object from GelSlim \cite{ian22} tactile readings by matching the sensed contact shape to simulated shapes in an object-specific, contrastively learned embedding. In robot manipulation, a more common scenario is handling previously unseen objects. Wang et al. \cite{wang21_1} proposed estimating object transformations by applying the Iterative Closest Point (ICP) algorithm to tactile-derived point clouds. NormalFlow \cite{huang24} uses tactile-derived normal maps to achieve very accurate pose tracking and performs well even on low-textured objects. \revised{Another common practice is to incorporate motion and geometric priors within a factor-graph framework, together with frame-to-frame relational constraints, to improve pose tracking accuracy. Sodhi et al. \cite{sodhi21} obtain these constraints from learned relative pose measurements between tactile image pairs. PatchGraph \cite{sodhi22} follows a similar formulation but estimates frame-to-frame transformations using ICP. Suresh et al. \cite{suresh21} instead derive them from implicit-surface contact constraint, where the implicit surface is obtained by integrating contact information over time.}

However, because tactile sensing captures only local information at each contact, tactile-based tracking methods for novel objects are often limited to short horizons and are prone to drift. Many approaches \cite{dikhale2022visuotactile, li25, murali2022active, suresh24, zhao2021novel, zhao23} address this by incorporating global information from vision, combining visual and tactile sensing to achieve robust tracking. In contrast, we challenge the common belief that tactile sensing is inherently restricted to local, short-horizon perception. Our work demonstrates that by incorporating robust loop detection methods, tactile sensing alone can overcome this limitation, enabling long-horizon object tracking with little drift.

\subsection{Object Geometry Reconstruction}
Besides object pose estimation, reconstructing dense 3D object geometry is important for many robotic tasks, including precise grasping and manipulation. It also plays a critical role beyond robotics, with applications in AR/VR, archaeological artifact reconstruction, biological structure analysis, and dental molding. The most common method for object reconstruction relies on vision. Structure-from-Motion (SfM) methods \cite{agarwal09, schoenberger16} recover scene geometry and camera poses from images alone by detecting and matching features across views, triangulating 3D points, and performing bundle adjustment. Visual SLAM methods \cite{kerl13, klein07, murartal15} estimate camera motion and build maps in real time. Moving beyond classical geometry, Neural Radiance Fields (NeRF) \cite{mildenhall20} and subsequent works \cite{li23, wang21} model scenes as neural fields of color and density, producing high-quality reconstructions. In contrast, 3D Gaussian Splatting (3DGS) \cite{kerbl23} represents scenes as sets of learnable 3D Gaussians, enabling fast novel-view synthesis and 3D reconstruction.

In uncontrolled environments or when dealing with small, occluded, or optically challenging objects, vision-based methods often struggle. Tactile sensing, in contrast, is unaffected by vision-related issues such as lighting variations and occlusions. Moreover, certain tactile sensors, such as GelSight \cite{yuan17}, can capture high-fidelity local geometric details that are difficult for vision to achieve. However, because touch is inherently local, tactile-dependent 3D reconstruction setup typically assumes a stationary object and a fully known touch pose from robot kinematics. A common setup is to use vision for global context and apply touch to probe occluded or unseen regions with known contact poses \cite{comi2024snapit, smith20203d, suresh2022shapemap, wang18, xu2023visual}. Touch-only reconstruction without vision likewise often assumes a known touch pose for each contact \cite{bauza2019tactile, comi2024touchsdf}. To improve efficiency, researchers have developed active exploration methods that target the most uncertain regions for both visual-tactile and tactile-only pipelines \cite{jamali2016active, rustler2022active, smith2021active, yi2016active}. While some approaches \cite{suresh24, zhao23} omit explicit contact-pose information, they typically rely on vision to provide global context.

In general, a truly ``in-the-wild'' touch-only 3D reconstruction algorithm that operates without assumptions or additional information remains missing. Early attempts \cite{huang24, lu23} are small-scale (fewer than $5$ loops), target mostly 2.5D geometries (planar or cylindrical surfaces), and are not robust. Our method shows that, even without known contact poses, vision, or any other external cues, tactile sensing alone can robustly produce high-fidelity dense reconstructions, including large targets such as tree trunks and low-texture objects.
\section{System Overview}

\begin{figure*}[htbp]
\centering
\includegraphics[width=0.98\linewidth]{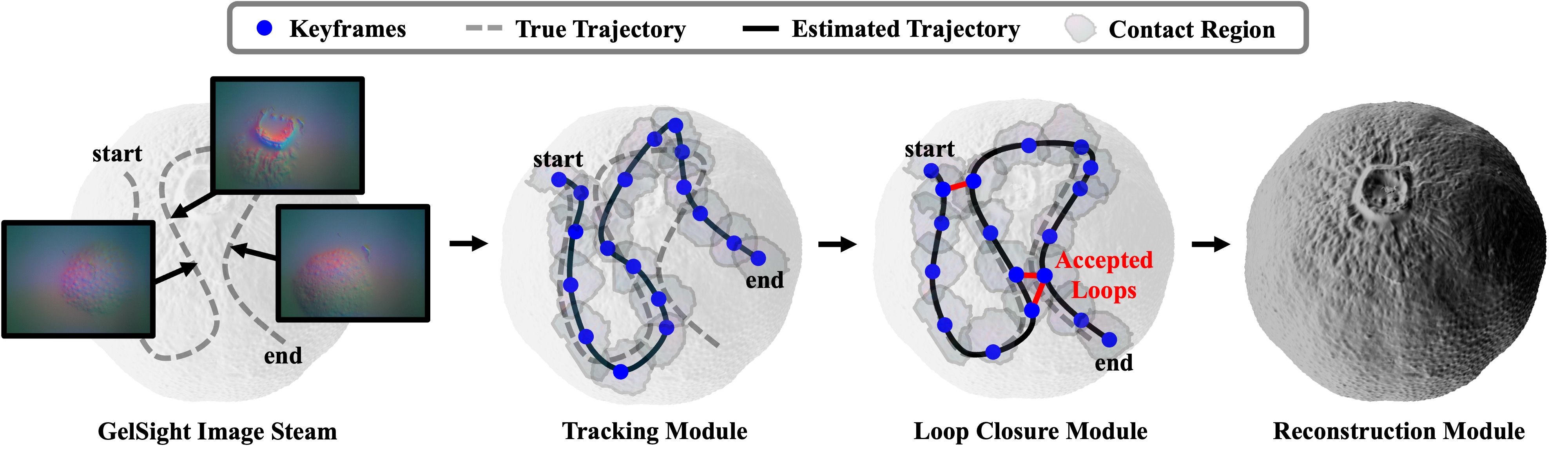}
\caption{GelSLAM pipeline: GelSight image streams are first processed by the tracking module to estimate object poses and select keyframes along the trajectory. Each new keyframe is passed to the loop closure module to identify revisits (loops). A globally consistent trajectory is then computed by optimizing a pose graph that combines tracking and loop information. The reconstruction module registers local tactile patches using the optimized poses and fuses them into the final 3D model.}
\label{fig:method_pipeline}
\amin{maybe making the contact region color to a more distinct color from the background will be better}
\end{figure*}

We aim to achieve real-time, accurate 6DoF long-horizon object tracking and high-fidelity object-level 3D reconstruction using only tactile input from a GelSight sensor. GelSight \cite{yuan17} is a vision-based tactile sensor that captures high-resolution images of local surface geometry by observing deformations on a soft elastomer. It provides dense but highly local geometric information. We consider a fully unconstrained setting where both the sensor and object are hand-held, can move freely, and may break and reestablish contact at any time. No proprioception, object model, or prior knowledge is assumed. The key challenge is that each GelSight image reveals only a small local patch, making it inherently difficult to relate tactile observations across space and time to recover global structure and long-horizon motion.

\subsection{Problem Statement}
% \wenzhen{you probably don't need it to be "real-time"}\joe{I updated it to long-horizon object tracking, which do need real-time.}

At each time step $i \in {0, 1, \dots, t}$, the GelSight sensor captures a tactile image $\mathbf{F}_i \in \mathbb{R}^{H \times W \times 3}$. Given the stream of tactile images ${\mathbf{F}_0, \mathbf{F}_1, \dots, \mathbf{F}_t}$, our goal is to estimate in real time the 6DoF trajectory ${\mathbf{T}_0, \mathbf{T}_1, \dots, \mathbf{T}_t}$, where each $\mathbf{T}_i \in SE(3)$ denotes the sensor pose relative to the object at time step $i$. While object pose tracking is typically described as estimating the object’s pose in the sensor frame, we instead adopt the sensor-relative formulation used in prior work such as~\cite{huang24}, as it aligns more naturally with the reconstruction task. The two are equivalent and easily convertible via matrix inversion. Without loss of generality, we set the initial pose as the reference: $\mathbf{T}_0 = I$. In addition to object pose tracking, we reconstruct the object as a mesh in two modes: online, which incrementally builds the shape in real time to guide users on where to scan next, and offline, which takes a recorded tactile video as input and reconstructs a high-fidelity 3D model.

%\note{Figure 4: A pipeline diagram showing the role of each module and how they are related.}
\subsection{Three Modules: Tracking, Loop Closure, Reconstruction}

GelSLAM (Fig. \ref{fig:method_pipeline}) consists of three modules, implemented to run as separate processes in parallel. \textbf{The tracking module} (Section \ref{sec:tracking}) estimates relative poses between temporally nearby frames, selects keyframes, and detects tracking failures caused by contact loss or rapid motion, triggering a new tracking session. However, the tracking module alone suffers from drift and cannot recover spatial relationships between disconnected tracking sessions. \textbf{The loop closure module} (Section \ref{sec:loop_closing}) addresses this by detecting revisited contacts between keyframes---even across long gaps in time or across separate tracking sessions---estimating their relative poses, and performing pose graph optimization to correct drift, relocalize, and ensure global pose consistency. \textbf{The reconstruction module} (Section \ref{sec:reconstruction}) fuses local surface patches into a global 3D mesh using the optimized poses. The reconstruction module can be disabled if only long-horizon pose tracking is required.
% \wenzhen{It's hard to understand the purpose of these three modules.}\joe{better now?}
\subsection{Key Insight: Differential Representations}
The key insight of GelSLAM is to estimate spatial transformations between GelSight images using differential representations of the surface: normal maps and curvature maps. Prior works \cite{li14} \cite{lu23} \cite{wang21_1} \cite{zhao23} often rely on point clouds or height maps. Because tactile contact occurs only at the very surface, the resulting height map is always shallow and lacks variation, with all height values remaining close to zero. Differential representations are unaffected by this limitation, as the first-order derivatives can still vary significantly and capture fine surface textures. This representation also aligns with GelSight’s sensing principle, which directly captures surface normals via photometric stereo, whereas point clouds require integrating normals, leading to accumulated noise.

%However, tactile contact only produces shallow deformations, resulting in flat, low-feature point clouds that are poorly suited for estimating spatial transformations. In contrast, differential representations retain fine surface textures. For example, a flat but textured surface like fabric may appear featureless in a point cloud yet reveal rich structure in a normal map. 
\begin{comment}
\joe{What is the definition of differential representation?}
\joe{Unnatural flow of differential representaitons. Point cloud is not the natural representation. Maybe try vision-based tactile sensors. How is the representation relavant to the structure.}
\joe{Not deficiency that tactile geometry is shallow. It is that zero-order reresentation doens't make clear features.}
\joe{Put back the old reason.}
\end{comment}

\begin{comment}
\begin{figure*}[htbp]
\centering
\includegraphics[width=0.98\linewidth]{Figures/key_insight.jpg}
\caption{Difference between the height, normal, and curvature maps at each object rotation and the reference map at $0^\circ$. As the object deviates from the reference pose, normal and curvature maps change smoothly with clear minima at the reference, while the height map changes erratically. This implies that pose estimation using height maps is less stable, while normal and curvature maps enable more robust and accurate estimation.\joe{We should keep it, but both Wenzhen and Michael find it unclear.}}
\label{fig:key_insight}
\end{figure*}
\end{comment}
\joe{I removed the key insight figure, but I want to think of a way to put it back.}

%\note{Figure 5: Horizontal axis: pose, vertical axis: loss. Above the loss function, show the tactile image, the height map, the normal map, and the curvature map.}
%Fig. \ref{fig:key_insight} compares height, normal, and curvature maps across frames with varying object poses (rotations). As the object deviates from the reference pose ($\text{rotation}=0^\circ$), normal and curvature maps change smoothly with clear minima at the reference, while height map differences remain noisy. This shows that differential representations offer more stable and reliable signals for estimating spatial transformations.

\subsubsection{Surface Normal Map}\label{sec:surface_normal_map}

GelSight \cite{yuan17} and related vision-based tactile sensors \cite{lambeta24} capture fine surface geometry by imaging deformations of a soft elastomer under contact. These sensors estimate surface normals at high spatial resolution using photometric stereo \cite{johnson11}. To recover normals from GelSight images, we adopt the data-driven approach from \cite{wang21_1}, widely used in recent works \cite{lu23, zhao23}. We denote the resulting surface normal map for frame $i$ as $\mathbf{I}_i \in \mathbb{R}^{H \times W \times 3}$, where each pixel stores a unit normal vector. We collect $50$ images by pressing a $6.31$ mm diameter metal ball against different regions of the sensor. We manually label the circular contact areas and analytically compute the ground truth surface gradients. A three-layer MLP (5–32–32–32–2) is trained to predict surface gradients $(g_u, g_v)$ from each pixel using a 5D input consisting of RGB color values and pixel coordinates $(u, v)$. At test time, the surface normal at each pixel is computed by normalizing the predicted gradient:
$$
\hat{\mathbf{n}} = \frac{\mathbf{n}}{\|\mathbf{n}\|}, \quad \text{where } \mathbf{n} = \begin{bmatrix} g_u & g_v & -1 \end{bmatrix}^\intercal.
$$

For auxiliary purposes, the height map $\mathbf{H}_i \in \mathbb{R}^{H \times W}$ is computed by integrating the predicted surface gradients using a 2D fast Poisson solver \cite{yuan17}. \revised{Specifically, the solver formulates gradient integration as a least-squares optimization problem that solves for the height map whose spatial derivatives best match the predicted gradients.} The contact mask $\mathbf{C}_i \in \{0,1\}^{H \times W}$ is then derived by thresholding the height map and the RGB value changes of each pixel in the tactile image. The height map can be converted into the point cloud representation by placing each pixel at its image location with the height value as its z-coordinate. The tactile-derived mesh can then be obtained by triangulating neighboring points in the point cloud according to the pixel grid structure.

% \amin{is there a reason why contact mask is not derived from the normals or rgb?}
% \joe{Edited.}
% \amin{I am curious whether the rgb was not enough for the contact mask.}\joe{I tried, not enough because of the shadows.}
\subsubsection{Surface Curvature Map}
\begin{comment}
While surface normal map have proven effective for frame-to-frame object tracking \cite{huang24}, they are unsuitable for appearance-based feature extraction methods such as SIFT \cite{lowe04}. Since surface normals rotate with the object, they lack rotational invariance, violating the fundamental assumption of these feature extraction methods that local appearance is spatially invariant.

Therefore, we introduce the curvature map. It is spatially invariant, meaning the surface curvature at any point remains unchanged under object transformations. Mathematically, surface curvature corresponds to the divergence of the normal field. In practice, we approximate it as the Laplacian of the surface height map, which we compute directly from the predicted surface gradients $(g_u, g_v)$. Specifically, we calculate $\partial g_u / \partial u + \partial g_v /\partial v$ at each pixel to obtain a dense scalar map that reflects local curvature. To remove high-frequency noise introduced by differentiation, we apply a $7 \times 7$ Gaussian filter, yielding a robust representation suitable for feature matching.
\end{comment}
Mathematically, surface curvature corresponds to the divergence of the surface normal field. We approximate it as the Laplacian of the surface height map, computed directly from the predicted gradients $(g_u, g_v)$ as $\partial g_u / \partial u + \partial g_v / \partial v$ at each pixel. To suppress high-frequency noise introduced by differentiation, we apply a $7 \times 7$ Gaussian filter. The resulting curvature map is a dense scalar field $\mathbf{L}_i \in \mathbb{R}^{H \times W}$. It is invariant to rigid transformations, meaning the curvature at each point remains unchanged under object translation or rotation, making it well suited for feature extraction methods such as SIFT \cite{lowe04}. In contrast, normal maps are not rotation invariant, as surface normals rotate with the object. Curvature also serves as a measure of local saliency, highlighting regions with rich texture and higher importance.

\joe{Maybe we should put the notation of normal map, height map, contact mask, and curvature map together at the beginning of the section.}

\begin{comment}
\begin{itemize}
    \item State the problem with the constraints: "in-the-wild", no proprioception, hand-held, can break contact. 
    \item Problem formulation: Tactile images.
    \item System overview: Tracking, Loop-closure, and Reconstruction.
    \item Differential Representations: Surface normal estimation and curvature estimation.
    \item Figure 1: System Overview, Example Pose Graph. Figure 2: Comparison between Differential Representation and Height map.
\end{itemize}
\end{comment}

\section{Method} \label{method}
In this section, we explain in detail the three core modules (Fig. \ref{fig:method_pipeline}) of the GelSLAM algorithm: the tracking module, the loop closure module, and the reconstruction module.

\subsection{The Tracking Module} \label{sec:tracking}

The tracking module performs short-horizon object pose tracking in a sequential manner. It maintains a sparse set of keyframes $\mathbb{K} = \{k_1, k_2, \dots, k_n\}$, where each $k_i$ indicates that frame $k_i$ (by its original index) is selected as the $i$-th keyframe. These keyframes act as anchors for estimating and composing relative poses. Specifically, the module estimates relative poses between consecutive keyframes and between each incoming frame and its most recent keyframe. These poses can be composed to compute the transformation between any pair of frames in the same tracking session. This keyframe-based strategy, commonly used in SLAM systems \cite{murartal15}, restricts loop closure detection and pose optimization to this small keyframe set, greatly improving computational efficiency. It also helps reduce drift by avoiding dense frame-to-frame composition.

We use NormalFlow \cite{huang24} for all pose estimations, whether between keyframes or between a frame and its most recent keyframe. To ensure robustness, we introduce metrics to evaluate estimation quality and detect failures. The following subsections detail the NormalFlow method, failure detection, and the full tracking pipeline.

\joe{Dan said maybe change the order: (1) this is how we predict ego motion (2) here is how we select keyframes. However, I think it's fine like the old way.}

\subsubsection{Relative Pose Estimation with NormalFlow}
We describe the general setup of NormalFlow \cite{huang24}. Given a reference frame $i$ and a target frame $j$, NormalFlow estimates the transformation from frame $i$ to frame $j$ by aligning their normal maps $\mathbf{I}_i$ and $\mathbf{I}_j$. The resulting transformation is denoted as ${^j\hat{\mathbf{T}}_i} = (\mathbf{R}, \mathbf{t}) \in SE(3)$, where the hat on $\hat{\mathbf{T}}$ indicates that the pose is estimated by NormalFlow. The transformation $(\mathbf{R}, \mathbf{t})$ is obtained by minimizing:

\begin{equation} \label{eq:nf}
\sum_{(u,v) \in \overline{\mathbf{C}}} \left\| \mathbf{I}_j(\mathbf{W}(u,v; \mathbf{R}, \mathbf{t})) - \mathbf{R} \, \mathbf{I}_i(u,v) \right\|^2
\end{equation}
where $\overline{\mathbf{C}}$ is the shared contact region, and $\mathbf{W}$ is the warping function under the estimated transformation $(\mathbf{R}, \mathbf{t})$:
\begin{equation} \label{eq:warp}
\mathbf{W}(u,v;\mathbf{R}, \mathbf{t}) = 
\begin{pmatrix}
1 & 0 & 0 \\
0 & 1 & 0
\end{pmatrix} \cdot
\left( \mathbf{R} \cdot \mathbf{q}(u,v)+ \mathbf{t} \right)
\end{equation}
where $\mathbf{q}(u,v)=\begin{bmatrix}u & v & \mathbf{H}_i(u,v)\end{bmatrix}^\intercal$ is the 3D point on the object's surface associated with pixel (u,v) in the reference frame. The objective is optimized using Gauss-Newton; see \cite{huang24} for details. NormalFlow runs at $8$\,ms per frame. To ensure robust and deterministic alignment, we replace NormalFlow’s original random subsampling in the summation of Equation (1) (used to reduce runtime) with a fixed selection of a fixed number of highest-curvature pixels from the reference frame within the shared contact region $\overline{\mathbf{C}}$. In our experiments with the GelSight Mini sensor, we set this number to $3000$.

\subsubsection{NormalFlow Failure Detection}
NormalFlow can silently return incorrect pose estimates due to poor initialization, insufficient shared contact, or local minima, with no internal signal of failure. To address this, we introduce two metrics to detect when NormalFlow pose estimation has failed. These metrics are designed to be applicable across contact scenarios and emphasize alignment in high-curvature regions.

\textbf{Curvature Cosine Similarity (CCS): } Let $\mathbf{L}_i$ be the curvature map of the reference frame, and $\mathbf{L}_j' = \mathbf{L}_j(\mathbf{W}(u,v; \mathbf{R}, \mathbf{t}))$ be the curvature map of the target frame warped to the reference frame using the estimated transformation ${^j\hat{\mathbf{T}}_i} = (\mathbf{R}, \mathbf{t})$. If the NormalFlow estimate is correct, the curvature maps $\mathbf{L}_i$ and $\mathbf{L}_j'$ should align closely. We compute their cosine similarity over the shared contact region $\overline{\mathbf{C}}$:

$$
\text{CCS} = \frac{\langle \mathbf{L}_j', \mathbf{L}_i \rangle}{\|\mathbf{L}_j'\| \cdot \|\mathbf{L}_i\|}
$$
This metric emphasizes alignment in textured regions, as the dot product gives more weight to high-curvature pixels. It is normalized and thus comparable across different objects regardless of their overall texture level, and upper bounded by 1 when perfectly aligned.

\textbf{Shared Curvature Ratio (SCR): } While CCS evaluates alignment quality, it may yield a high score even when the shared contact region is small. To ensure sufficient spatial overlap, we introduce the Shared Curvature Ratio, which measures the proportion of the reference frame's contact region that overlaps with the target frame's, weighted by curvature.

$$
\text{SCR} = \frac{\sum_{(u,v)\in \overline{\mathbf{C}}} \mathbf{L}_i(u,v)}{\sum_{(u,v)\in \mathbf{C}_i} \mathbf{L}_i(u,v)}
$$

%NormalFlow is considered failed if $\text{CCS} < 0.7$ or $\text{SCR} < 0.3$, indicating poor alignment or insufficient overlap between frames.
NormalFlow is considered failed if CCS or SCR is below a predefined threshold, indicating poor alignment or insufficient overlap between frames.

\subsubsection{Pipeline and Keyframe Selection} \label{tracking:pipeline}
For each new frame $t$ with tactile image $\mathbf{F}_t$, the tracking module computes its geometric properties $\{\mathbf{I}_t, \mathbf{L}_t, \mathbf{H}_t, \mathbf{C}_t\}$ and estimates its pose relative to the most recent keyframe using NormalFlow, initialized by the previous pose estimate at $t{-}1$ to that keyframe. If the pose estimation with NormalFlow fails (low CCS or SCR), the previous frame \(t{-}1\) is set as a new keyframe and added to \(\mathbb{K}\). The estimated transformation between the new keyframe and its predecessor keyframe is recorded for use as a pose constraint in later pose graph optimization (Section \ref{sec:pose_graph}). Unlike \cite{huang24}, which runs NormalFlow twice per frame for keyframe selection, our method uses only a single run.
%\amin{I am not completely understanding the difference between the following scenario with the above}

By design, the latest keyframe is usually several frames behind the current frame because we defer keyframe updates until NormalFlow can no longer track reliably. In the extreme case where the latest keyframe is the previous frame $t{-}1$ and the NormalFlow estimate from $t{-}1$ to $t$ still fails, we declare tracking lost and initialize a new tracking session with frame $t$ as the first keyframe in the new tracking session. This typically results from contact loss or rapid motion.

\subsection{The Loop Closure Module} \label{sec:loop_closing}
While the tracking module provides short-horizon pose tracking, it suffers from drift and cannot relate frames across tracking sessions. The loop closure module addresses this by detecting revisits, referred to as loops, between the new keyframe and existing keyframes. A loop is defined as a pair of keyframes with overlapping contact regions and an estimated relative pose between them. Detecting such loops enables spatial alignment over long time intervals or across tracking sessions. To avoid exhaustive matching, we detect loops only against a subset of keyframes, called \textbf{the coverage set}. We then apply pose graph optimization using pairwise pose constraints from both tracking and loop closures to recover globally consistent keyframe poses and achieve accurate long-horizon pose tracking.

\begin{figure}[htbp]
\centering
\includegraphics[width=0.98\linewidth]{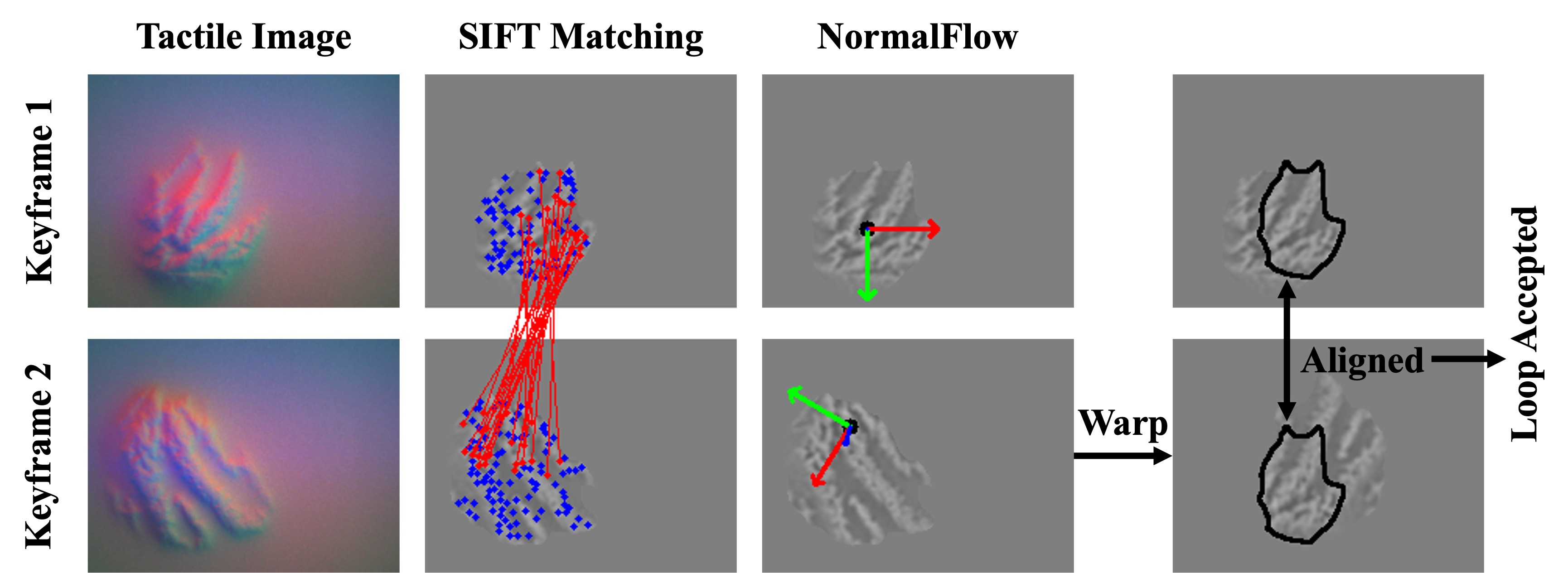}
\caption{Loop detection pipeline. SIFT features on curvature maps (second column) are first matched to estimate an initial relative transformation, which is then refined by NormalFlow (third column). If the warped curvature maps (fourth column) are well aligned and the shared contact region is sufficiently large, the loop is accepted. This verification is performed by NormalFlow’s failure detection process, which evaluates the CCS and SCR scores.}
\label{fig:loop_detection}
% \amin{having abc for the columns and caption can help matching them}
\end{figure}
%NormalFlow failure is detected by evaluating the alignment between the warped curvature maps (fourth column); if the difference is small (high CCS) and the shared contact region is large (high SCR), the loop is accepted.

\subsubsection{The Coverage Set of Keyframes} \label{sec:coverage_set}
\joe{Dan: This section feels like it glosses over the definition of coverage too much. How is coverage computed, and maintained over long periods of time? How is redundancy computed? by some overlap threshold? Given the specifics in the prev section this one feels like its is very hand-wavy}\joe{I fixed the paragraph based on Dan's siggestion.}
To reduce online computation, loops are detected only against a selected subset of keyframes, referred to as the coverage set $\mathbb{C} \subseteq \mathbb{K}$. The keyframes in $\mathbb{C}$, termed \textbf{the coverage keyframes}, are chosen to collectively span all scanned contact regions, each contributing non-redundant surface areas. As a result, if a new keyframe revisits a region previously scanned by any keyframe, it is likely to also overlap with a coverage keyframe. The coverage set $\mathbb{C}$ is built in a greedy manner. It is initialized as an empty set, and whenever a new keyframe is introduced, we evaluate how much new contact region the new keyframe contributes beyond what is already covered by $\mathbb{C}$. If this new contact area exceeds 0.2\,mm$^2$, the keyframe is added to the set. Afterward, we check whether any existing coverage keyframe has become redundant. A coverage keyframe is considered redundant if it does not contribute more than 0.2\,mm$^2$ of contact region that is not covered by the other keyframes in $\mathbb{C}$. Any such keyframes are pruned.
% \amin{So, am I right to say that the contact set is somehow the merged version of the K? where nearby points have only one representation?}
% \joe{I am not sure what you mean. The coverage set is a subset of K. But you can say that for any keyframe in K, it is covered by the union of the coverage set. The coverage set is not merged from K, but a subset of K.}

\subsubsection{Loop Detection}
%\note{Figure 6: The two-staged loop detection process. Showing SIFT matching then NormalFlow.}
When a new keyframe is selected by the tracking module, the loop closure module searches for loops by comparing it against keyframes in the coverage set using a two-stage process (Fig. \ref{fig:loop_detection}). First, SIFT features \cite{lowe04} are extracted from the curvature map of the new keyframe and matched to those from coverage keyframes. The curvature map is well suited for SIFT feature extraction and matching, as it satisfies the assumption of appearance invariance under spatial transformations. From the matched SIFT features, a planar 2D transformation is estimated using a least-squares method with RANSAC for outlier rejection. Pairs with more than $8$ inlier matches are retained as loop candidates. The first stage runs at 0.2\,ms per pair. In the second stage, NormalFlow estimates the full 6DoF relative pose for each candidate pair, initialized by the estimated 2D transformation from SIFT matching. This stage runs at an average of 8\,ms per pair on a standard CPU-only laptop.

Loops with high-quality NormalFlow estimates (high $\text{CCS}$ and high $\text{SCR}$) are accepted. When running online, loop detection for a keyframe may not finish before the next keyframe is added by the tracking module. In such cases, we skip loop detection on any intermediate keyframes and resume it only from the next available keyframe after the current loop detection completes.
% and effective due to limited roll and pitch from contact constraints

\subsubsection{Pose Graph Optimization}\label{sec:pose_graph}

To estimate globally consistent poses for all keyframes, we solve a pose graph optimization problem over pairwise pose constraints, which come from either consecutive keyframes in tracking or pairs from the detected loops:

\begin{equation}\label{eq:pose_graph}
\begin{aligned}
& \argmin_{\{\mathbf{T}_{k_1}, \dots, \mathbf{T}_{k_n}\}} 
\sum_{(i,j) \in \mathcal{E}} 
\left\| \mathbf{e}_{ij} \right\|^2_\mathbf{\Sigma} \\
& \quad \text{with} \quad 
\mathbf{e}_{ij} = \log\left( {^{k_j}\hat{\mathbf{T}}_{k_i}}^{-1} \cdot \mathbf{T}_{k_j}^{-1} \mathbf{T}_{k_i} \right)
\end{aligned}
\end{equation}
Here, \(\mathbf{T}_{k_1}, \dots, \mathbf{T}_{k_n} \in SE(3)\) are the keyframe poses to be estimated, and \(\mathcal{E}\) is the set of keyframe pairs \((i, j)\) for which a pairwise pose constraint is available. For each pairwise constraint, the pose error \(\mathbf{e}_{ij}\) is the logarithmic map of the difference between the predicted relative pose \(\mathbf{T}_{k_j}^{-1} \mathbf{T}_{k_i}\) and the NormalFlow-estimated relative pose \({^{k_j}\hat{\mathbf{T}}_{k_i}}\), yielding a 6D error vector in \(\mathfrak{se}(3)\). All constraints share a fixed covariance matrix \(\mathbf{\Sigma}\), as we find that using estimated covariances does not improve the performance in practice. We solve the resulting nonlinear least-squares problem using the Levenberg--Marquardt algorithm \cite{Henri23} implemented in GTSAM \cite{gtsam22}. Since every frame is associated with a keyframe, the poses of all frames can be recovered from the optimized keyframe poses.

\subsection{The Reconstruction Module} \label{sec:reconstruction}
%\note{Figure 7: Picking two objects, show the three version of reconstruction. (1) directly fusing (2) projection fusing (ours) (3) re-meshing}

\begin{figure}[htbp]
\centering
\includegraphics[width=0.85\linewidth]{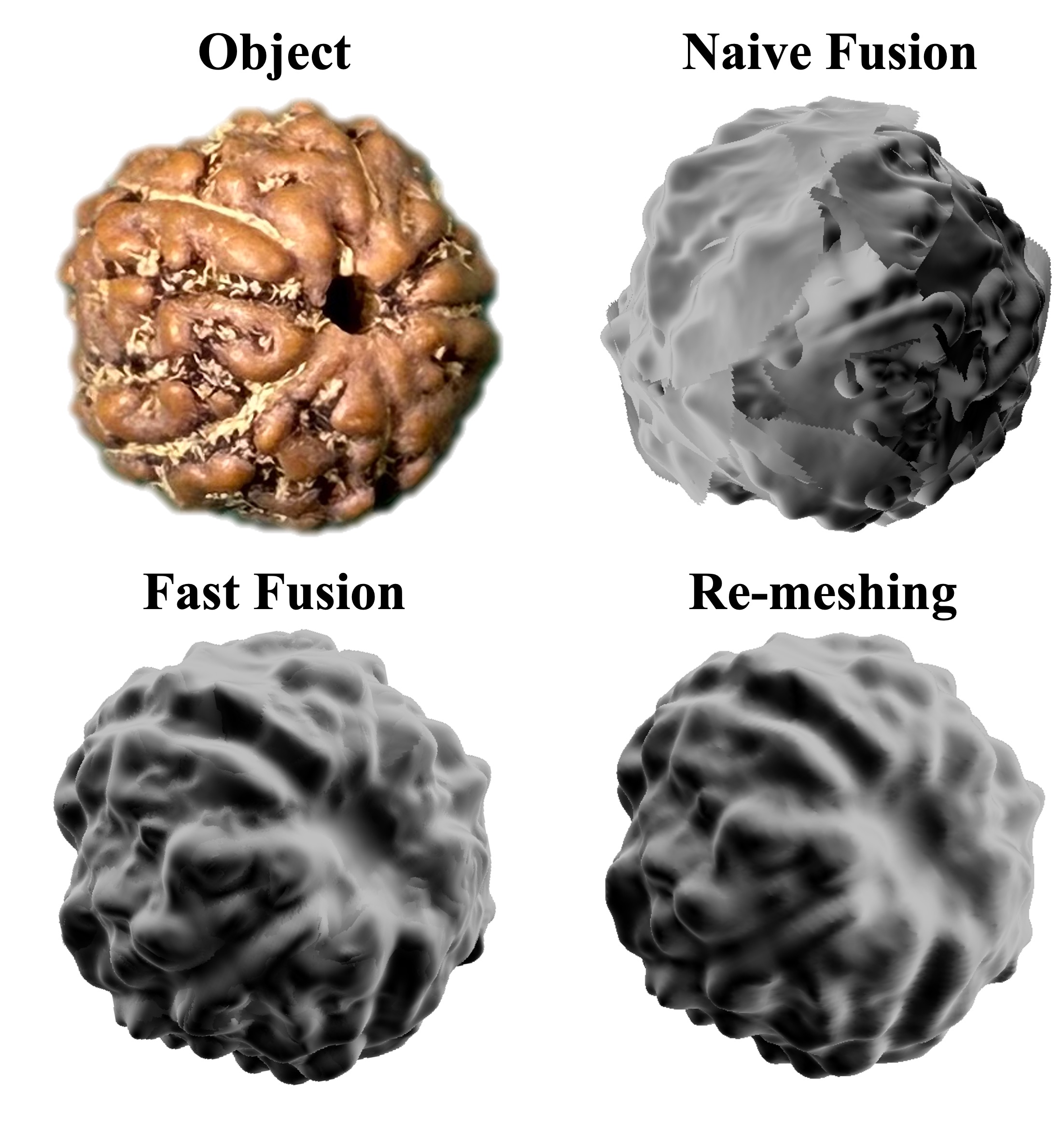}
\caption{Reconstruction pipeline. Naively pasting tactile meshes at their estimated poses can introduce artifacts (naive fusion). We address this with a fast fusion step, which can run in real time to provide online reconstruction feedback, followed by an offline re-meshing step to produce a watertight final model.}
\label{fig:reconstruction_method}
\end{figure}

Given the optimized keyframe poses from pose graph optimization, we reconstruct the object in two phases (Fig. \ref{fig:reconstruction_method}): a fast surface fusion step designed for online use, and a re-meshing step that produces a watertight reconstruction.

The fast fusion step creates a fused surface by registering and averaging point clouds from the coverage keyframes. This step is designed for efficient online use to provide real-time feedback and help guide user scanning. Naively transforming and merging individual point clouds often leads to artifacts (upper right in Fig. \ref{fig:reconstruction_method}), particularly near the borders of contact regions, where depth distortions are common. To reduce these issues, we average overlapping regions across the coverage keyframes. For each pixel in a coverage keyframe, we reproject it into overlapping keyframes in the coverage set. If the reprojected pixel falls within the other keyframe’s contact region, we retrieve its corresponding 3D point. The final point location is then computed as a weighted average of these 3D points. Each point is weighted using a sigmoid function of its distance to the contact boundary in the keyframe it maps to, down-weighting unreliable border regions. We triangulate the resulting averaged point cloud for each coverage keyframe and merge them into a global mesh. The result is visually coherent and fast to compute, though not watertight.

The second phase re-meshes this fused surface to produce a watertight, high-fidelity mesh. We perform re-meshing using the Poisson Surface Reconstruction method \cite{kazhdan2013screened}. Since the goal of reconstruction is to obtain an accurate object shape rather than support real-time operation, this phase is treated as an offline process.
% \amin{So I mostly used the code for Poisson surface remeshing to make it watertight. For large objects, I could get out of memory running the script, so I used Blender for them.}
%We use the implementation provided by Open3D \cite{Zhou2018}. 

\begin{comment}
\begin{itemize}
    \item Explain the two types of reconstruction, online and offline.
    \item Given the solved keyframe poses from pose graph optimization, we apply a two-staged 3D reconstruction.
    \item Reconstruction is based on averaging the coverage set on their solved poses. In practice, using all keyframes is similar to using only the coverage set. We use this because it is fast, since the pose of a keyframe can change over time (due to detected loops), implicit representation for mesh merging needs to happen every step.
    \item Explain the mesh merging in detail.
    \item Explain the Re-meshing using Blender.
    \item Figure: Side-by-side, simple reconstruction, averaging reconstruction, re-meshed.
\end{itemize}
\end{comment}

\revised{\subsection{Summary of Contributions}
In summary, our main contributions are the algorithmic design of novel SLAM components tailored to tactile sensing and their integration into three modules that form a well-engineered tactile SLAM system. The proposed components include NormalFlow failure detection, keyframe selection, and loop detection. Conventional SLAM components do not transfer well to tactile sensing due to the fundamentally different challenges it poses compared to vision and other sensing modalities. Accordingly, we incorporate our key insight on differential representations into the design of these components. For example, standard loop detection methods based on matching SIFT features assume rotational and translational invariance, which tactile images do not exhibit. Instead, we match SIFT features of the curvature map, which can be accurately estimated from tactile data, are invariant to rigid transformations, and remain informative even when tactile point clouds are nearly planar.}

\revised{While GelSLAM is built on NormalFlow \cite{huang24}, NormalFlow itself can not detect its own tracking failures or perform relocalization after contact is lost, and remains a purely local method. Our novel components are key to enabling robust, real-time, and large-scale tracking and reconstruction. With these components, GelSLAM robustly detects thousands of loops across tens of thousands of frames, whereas NormalFlow and other existing approaches, such as Tac2Structure \cite{lu23}, scale only to hundreds of frames and suffer from frequent false loop detections.}

\section{Experiment: Long-horizon Tracking }
In this section, we evaluate the performance of GelSLAM on real-time, long-horizon 6DoF object pose tracking.
% , comparing it against baseline methods and ablations across a wide variety of objects. We also report relevant system characteristics such as runtime and component-level statistics. 
Our experiments use the GelSight Mini tactile sensor, configured to a resized resolution of $320 \times 240$. The sensor has an effective sensing area of $20\,\text{mm} \times 15\,\text{mm}$, a frame rate of 25\,Hz, and a maximum indentation depth of about $2$\,mm. The system is evaluated on an AMD Ryzen 7 PRO 7840U CPU with 8 cores at 3.3\,GHz. GelSLAM is implemented in ROS2 \cite{macenski22}.

\joe{Double check on the 2mm indentation.}

\joe{Explain that the evaluation is mainly about how to correct drifts.}

\subsection{Data Collection}
We construct an object pose tracking dataset similar to \cite{huang24}, but designed to evaluate long-horizon tracking. The dataset in \cite{huang24} contains only short-range motions and requires only a single keyframe, making it insufficient for our evaluation. 

\begin{figure}[t]
\centering
\includegraphics[width=0.7\linewidth]{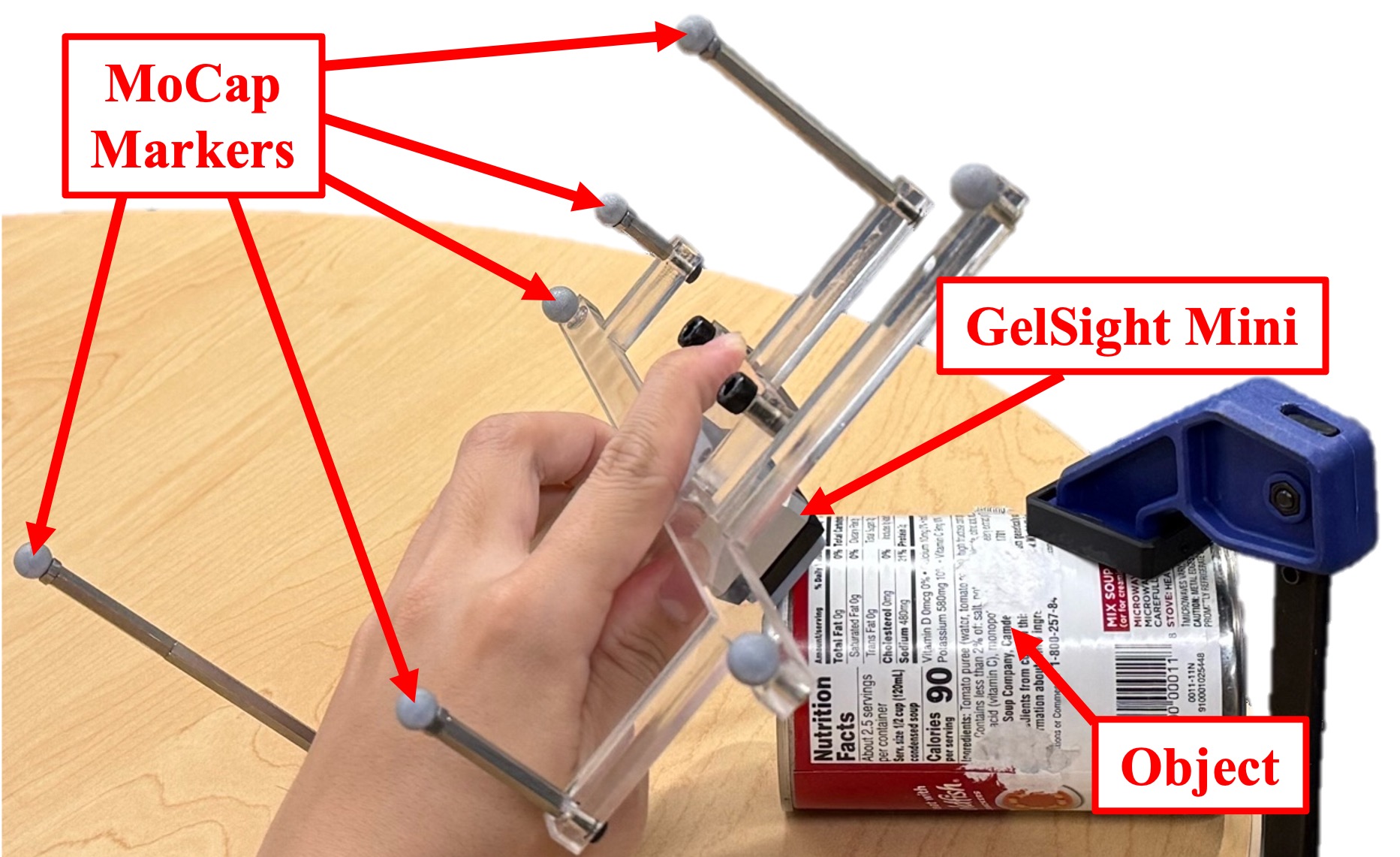}
\caption{Tracking data collection setup: the object is clamped to the table, and the GelSight sensor is tracked using MoCap.}
\label{fig:setup}
\end{figure}

\begin{figure}[t]
\centering
\includegraphics[width=0.98\linewidth]{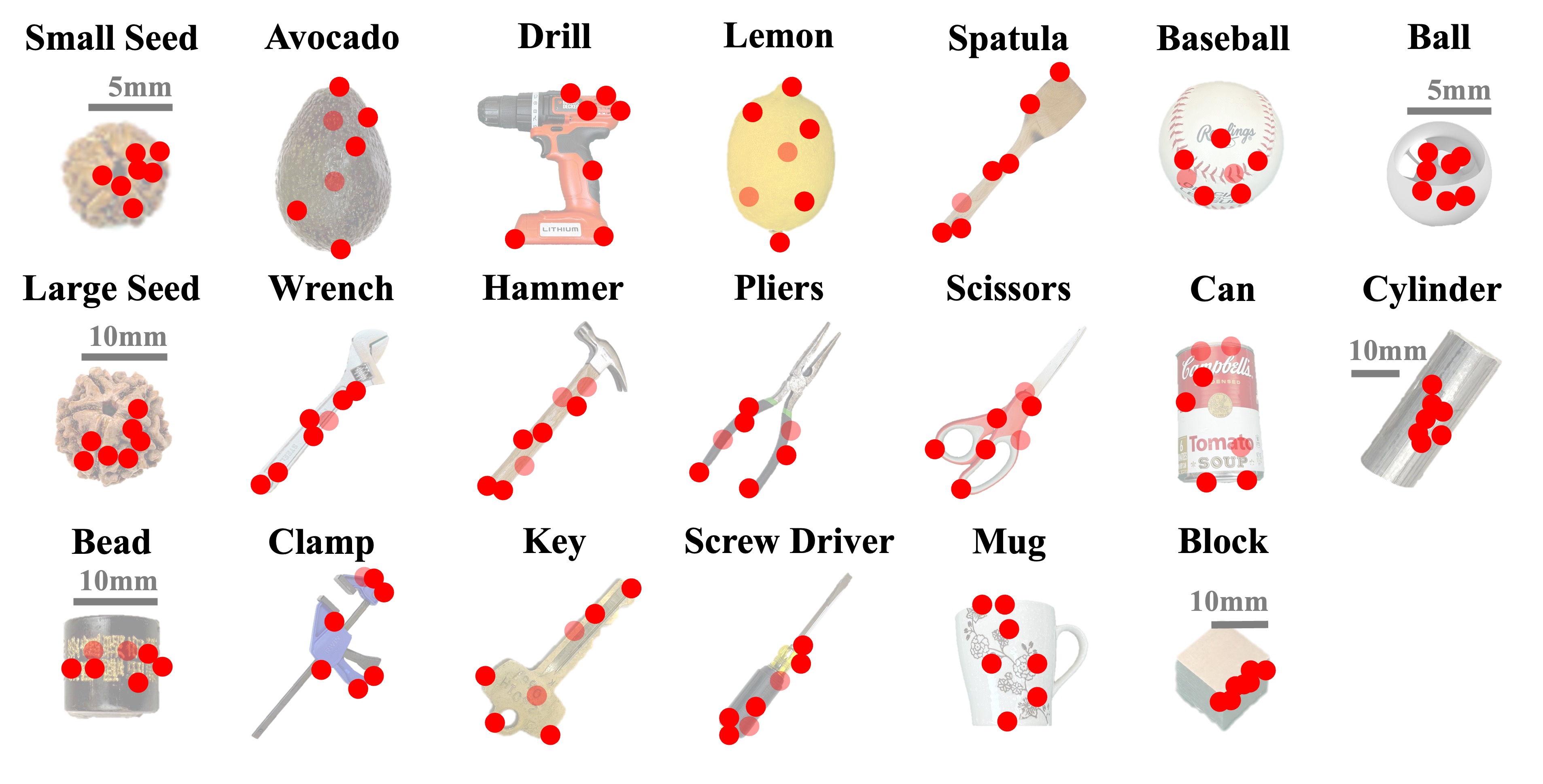}
\caption{Initial contact locations for the seven episodes per object in the tracking experiment, manually labeled. All objects are rescaled for visualization; actual sizes vary, and scales are shown only for uncommon objects.}
\label{fig:contact_locations}
\end{figure}

We collect our dataset using 20 objects across three categories: $14$ common items, $3$ small textured objects, and $3$ simple geometric primitives. Eleven of the twelve objects from \cite{huang24} are included, excluding the Table. During data collection, each object is rigidly clamped to a tabletop, and the GelSight Mini sensor is mounted on a mobile platform tracked by an OptiTrack motion capture (MoCap) system to provide ground-truth 6DoF poses (Fig. \ref{fig:setup}). For each object, we record seven episodes, with initial contact poses manually selected at random (Fig. \ref{fig:contact_locations}), yielding a total of $140$ episodes. The sensor is moved across a wide area. Continuous contact is maintained throughout, as baselines without loop closure can not recover once contact is lost. On average, each episode lasts $21$ seconds and contains $523$ frames. Table~\ref{tab:average_displacement} shows the average cumulative 6DoF motion across all episodes, which is substantially higher than in \cite{huang24}.
% \amin{1080 degrees is cumulative? I didn't completely understand the connection between the text and table. Is the 1080 and 93 the total xyz of the table?}\joe{Oh, they are basically the length of the vector. I removed that for clarity.}

\begin{table}[ht]
\centering
\begin{tabular}{|c|c|c|c|c|c|}
\hline 
$\theta_x(^{\circ})$ & $\theta_y(^{\circ})$ & $\theta_z(^{\circ})$ & x(mm) & y(mm) & z(mm) \\ 
\hline 
\hline 
452 & 516 & 617 & 44.7 & 51.2 & 40.2\\
\hline 
\end{tabular}
\caption{Average cumulative 6DoF motion per episode.}
\amin{is it the template to have the table caption under the table?}\joe{what do you mean?}
\label{tab:average_displacement}
\end{table}

\subsection{System Characteristics and Performance}
We run GelSLAM online across the 140 episodes of data. For NormalFlow failure detection, we adopt a conservative threshold: the estimate is rejected if $\text{CCS} < 0.85$ or $\text{SCR} < 0.3$. This threshold is fixed across all episodes. Table \ref{tab:tracking_statistics} summarizes GelSLAM statistics across all episodes. GelSLAM constructs keyframes much more frequently than visual SLAM approaches \cite{murartal15}, generating one every $7.7$ frames. This is necessary because tactile sensing is highly local, and even frames captured just a short time apart can have little or no overlap. Although many keyframes are introduced, the number of coverage keyframes remains low, as they represent only uniquely scanned surface regions. At the same time, the large number of keyframe pairs is efficiently reduced by SIFT matching to a manageable set of loop candidates. The high ratio of accepted loops among these candidates shows that the filtering preserves inliers while discarding most outliers. 

\begin{table}[ht]
\centering
\begin{tabular}{|c|c|c|c|c|c|c|}  %<-- change here
\hline 
& Frames & Keyframes & \makecell{Coverage \\ Keyframes} & \makecell{Loop \\ Candidates} & Loops \\ 
\hline 
\hline 
Mean & 523 & 68 & 21 & 65 & 34   \\ 
\hline 
Std & 50 & 59 & 14  & 78 & 36  \\ 
\hline 
\end{tabular}
\caption{GelSLAM statistics for long-horizon pose tracking (mean and standard deviation across the 140 episodes).}
\label{tab:tracking_statistics}
\end{table}

Table \ref{tab:tracking_runtime} presents the runtime breakdown of each GelSLAM module during online operation. The tracking module runs faster than the GelSight’s 25\,Hz frame rate. The method ``compute geometric properties" refers to the step that takes a GelSight image $\mathbf{F}_t$ and computes $\{\mathbf{I}_t, \mathbf{L}_t, \mathbf{H}_t, \mathbf{C}_t\}$. In the loop closure module, loop detection is the most time-consuming step, \revised{with a runtime ($12.6 \pm 25.2$ ms) that depends on the number of loop candidates. Since this module is triggered only when a new keyframe is inserted, on average every $310$ ms, its runtime is well within the time budget for online operation.}

\begin{figure}[t]
\centering
\includegraphics[width=0.98\linewidth]{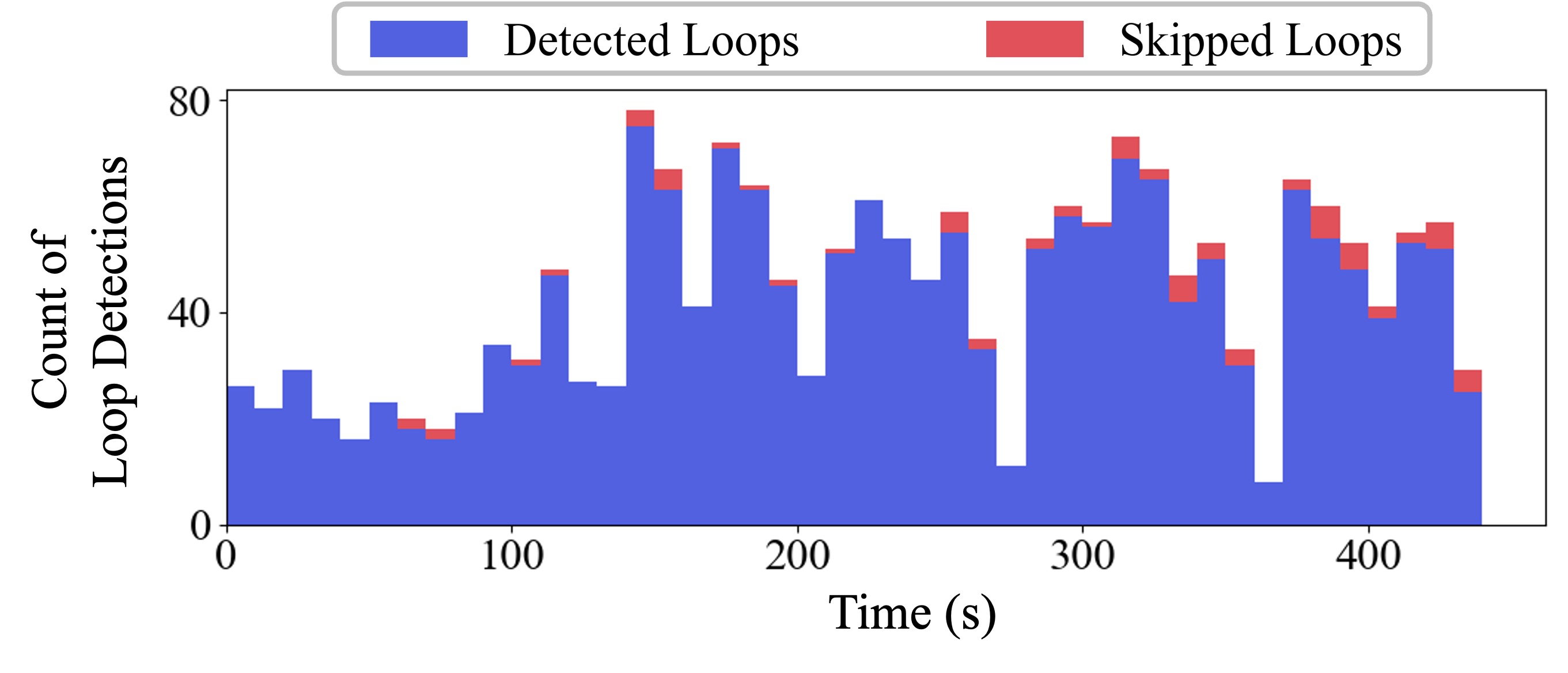}
\caption{\revised{Detected and skipped loop closures over time during online execution. Fewer than 5\% of loops are skipped, even in a very long episode.}}
\label{fig:loop_detection_frequency}
\end{figure}

\revised{As the episode length increases, loop detection runtime can increase, potentially causing some keyframes to be skipped for loop closure during online execution. To evaluate this effect in long episodes, we analyze a seven-minute episode tracking the Hammer. Fig. \ref{fig:loop_detection_frequency} shows the number of detected and skipped loops over time. No loops are skipped during the first minute, which already exceeds the duration of most manipulation tasks. Over the full seven-minute episode, fewer than 5\% of loops are skipped. Overall, the impact on online operation, due to skipped loop closures, is negligible for realistic task durations and remains small even in unusually long episodes.}

\begin{table}[ht]
\renewcommand{\arraystretch}{1.2}
\centering
\begin{tabular}{|c|c|c|c|} 
\hline
\textbf{Module} & \textbf{Method} & \textbf{Mean (ms)} & \textbf{Std (ms)} \\
\hline
\hline
\multirow{3}{*}{Tracking} 
    & Compute Geometric Properties & 19.2 & 2.5\\
    & NormalFlow Tracking    & 8.0 & 6.0 \\
    \cline{2-4}
    & Total    & 27.9 & 6.8 \\
\hline
\hline
\multirow{4}{*}{\makecell{Loop \\ Closure}} 
    & Loop Detection    & 12.6 & 25.2 \\
    & Pose Graph Optimization & 1.4 & 2.3 \\
    & Coverage Set Update & 10.4 & 15.9 \\
    \cline{2-4}
    & Total    & 33.3 & 33.3 \\
\hline
\end{tabular}
\caption{Runtime breakdown of GelSLAM modules, where loop closure runtime is measured per keyframe and tracking runtime is measured per frame.}
\label{tab:tracking_runtime}
\end{table}

%\note{Figure 9: runtime dotted plot of all trials (for loop closure). Horizontal axis is time, and vertical axis is runtime.}
%\joe{Is Figure 9 redundant compared to Table III? I want to show that the loop closure module's runtime increases only sublinearly over time.}

\subsection{Baseline and Ablation Methods}
We compare the online performance of GelSLAM (referred to as \textbf{GS-Online}) against five baselines: four frame-to-frame tracking methods \cite{chen91, wei19, huang24, radu09} and one full SLAM method (Tac2Structure \cite{lu23}). All baselines are run offline, as some can not keep up with GelSight’s frame rate. For baseline methods that require a point cloud, we compute it using the approach described in Section \ref{sec:surface_normal_map} and remove points outside the computed contact mask. Below, we briefly describe each baseline:
\begin{itemize}
    \item \textbf{ICP \cite{chen91}:} Refers to the Point-to-Plane ICP method, a point cloud registration approach used for frame-to-frame tactile tracking in \cite{sodhi22, wang21_1}. We use the implementation provided by Open3D \cite{Zhou2018}.
    \item \textbf{FilterReg \cite{wei19}:} A probabilistic point cloud registration method, used for frame-to-frame tactile-based tracking in \cite{bauza23}. We use the implementation from ProbReg \cite{probreg2019}.
    \item \textbf{FPFH+RI \cite{radu09}:} Refers to FPFH + RANSAC + ICP, this method combines feature-based matching with point cloud registration. It is used for frame-to-frame tactile-based tracking in \cite{lu23}. We use the implementation from Open3D \cite{Zhou2018}.
    \item \textbf{NF \cite{huang24}:} Refers to NormalFlow, a state-of-the-art method for tactile-based tracking. This baseline performs frame-to-frame tracking without keyframe selection. We use the implementation released with the original paper.
    \item \textbf{Tac2Structure \cite{lu23}:} A tactile-based full SLAM pipeline. It uses FPFH+RI for frame-to-frame tracking and CNN-based feature cosine similarity for loop detection. We re-implement it based on the original paper.
\end{itemize}

To assess the contribution of each component in GelSLAM, we compare against six ablation variants. These are run offline to avoid asynchronous behavior and better isolate the effect of each component. The ablation methods are:
\begin{itemize}
    \item \textbf{NF-OrigKF:} NormalFlow tracking with the original keyframe selection from the NormalFlow paper \cite{huang24}.
    \item \textbf{NF-KF:} The full GelSLAM tracking module, consisting of NormalFlow tracking with our keyframe selection method.
    \item \textbf{GS-OnlySIFT:} GelSLAM's tracking module with loop detection based solely on SIFT matching on the curvature map, without the second-stage NormalFlow refinement.
    \item \textbf{GS-OnlyNF:} GelSLAM's tracking module with loop detection performed directly using NormalFlow, without the first-stage SIFT-based candidate selection.
    \item \textbf{GS-ImageSIFT:} GelSLAM's tracking module with the two-stage loop detection, where the first-stage SIFT matching on the curvature map is replaced with SIFT matching on the raw GelSight image.
    \item \textbf{GS-Offline:} The full GelSLAM pipeline running offline. Unlike the online version, it includes all detected loops, some of which may be skipped in GS-Online.
\end{itemize}

\begin{comment}
Explain the baseline methods: 
\begin{itemize}
    \item Tracking without loop-closing: ICP, FilterReg, FPFH-R, NormalFlow.
    \item Full SLAM: Tac2Structure
    \item Ablations: NormalFlow + keyframe, NormalFlow + old keyframes, GelSLAM - SIFT Matching, GELSLAM - NormalFlow refinement, GelSLAM replacing SIFT matching to raw SIFT matching, GelSLAM (offline), GelSLAM (online). Note that all of our approach uses CCS and SCR with a special threshold.
\end{itemize}
Note that all ablations are done offline. The only difference between online and offline is: it is synchronous and it can involve more loops. Since ablation is to compare the contribution of each component, we run it offline to make sure they are compared from the same basis.
\end{comment}

\subsection{Long-horizon Tracking Results}

\begin{table}[ht]
\renewcommand{\arraystretch}{1.2}
\centering
\begin{tabular}{|@{\hskip 4pt}c@{\hskip 4pt}|@{\hskip 4pt}c@{\hskip 4pt}|
                @{\hskip 4pt}c@{\hskip 4pt}|@{\hskip 4pt}c@{\hskip 4pt}|
                @{\hskip 4pt}c@{\hskip 4pt}|@{\hskip 4pt}c@{\hskip 4pt}|
                @{\hskip 4pt}c@{\hskip 4pt}|}
\hline 
Method & $\theta_x(^{\circ})$ & $\theta_y(^{\circ})$ & $\theta_z(^{\circ})$ & x(mm) & y(mm) & z(mm) \\ 
\hline 
\hline 
ICP & 13.0 & 13.9 & 23.9 & 8.14 & 9.94 & 6.73 \\ 
\hline
FilterReg & 13.7 & 13.7 & 19.2 & 3.87 & 3.49&  2.61\\ 
\hline 
FPFH+RI & 37.8 & 38.7 & 36.3 & 90.6 & 134.1 & 108.5 \\ 
\hline 
NF & 7.11 & 7.53 & 7.63 & 1.13 & 1.21 & 0.92 \\ 
\hline 
Tac2Structure & 37.7 & 38.7 & 36.3 & 90.6 & 134.0 & 108.0 \\ 
\hline 
\hline
NF-OrigKF & 6.74 & 7.18 & 7.12 & 1.35& 1.38 & 0.91 \\ 
\hline 
NF-KF & 6.55 & 7.39 & 7.11 & 1.26 & 1.24 & 0.90 \\ 
\hline 
GS-OnlySIFT & 8.13 & 8.37 & 4.17 & \bf{1.04} & \bf{1.02} & 0.89 \\ 
\hline 
GS-OnlyNF & 5.61 & 6.28 & 6.95 & 1.53 & 1.62 & 0.77 \\ 
\hline 
GS-ImageSIFT & 5.49 & 6.13 & 6.00 & 1.16 & 1.14 & 0.83 \\ 
\hline 
GS-Offline & \bf{3.98} & \bf{4.29} & \bf{3.41} & \bf{0.98} & \bf{0.94} & \bf{0.71} \\ 
\hline 
\makecell{GS-Online\\(Ours)} & \bf{4.06} & \bf{4.38} & \bf{3.57} & \bf{1.00} & \bf{0.96} & \bf{0.72} \\ 
\hline 
\end{tabular}
\caption{Average 6DoF tracking MAE for all methods across all objects, excluding the Ball and Cylinder due to their perfectly smooth and symmetrical geometry.\joe{Is it clear what the bolded number means?}}
\label{tab:tracking_result}
\end{table}

\begin{figure*}[htbp]
\centering
\includegraphics[width=0.95\linewidth]{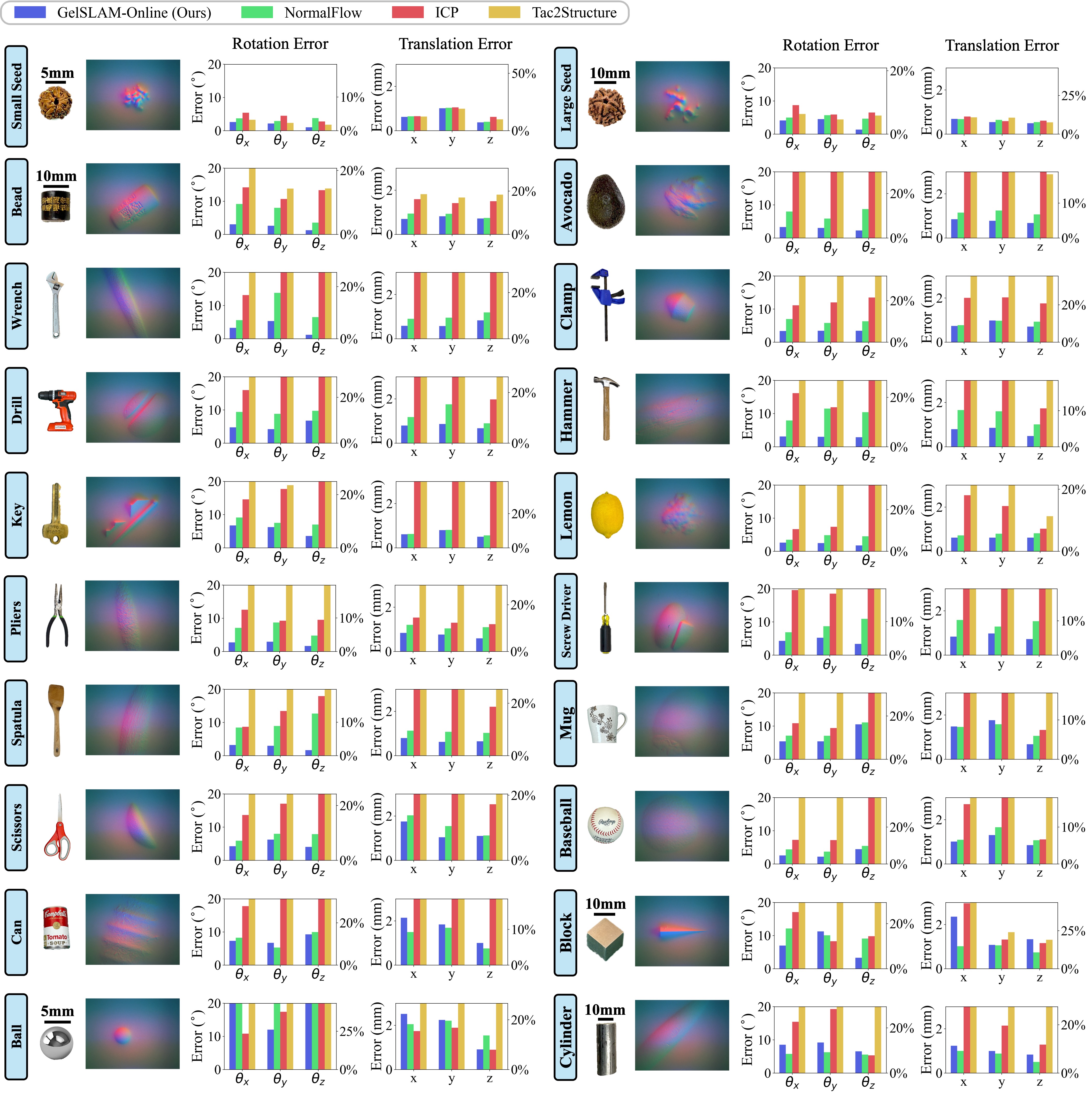}
\caption{Tracking results for the 20 objects, with 7 episodes per object (140 total). For each object: \textbf{[left]} the object image and an example tactile image; \textbf{[right]} the 6DoF mean absolute error (MAE), where the left y-axis shows the absolute error and the right y-axis shows the percentage error relative to the object's motion range. Errors exceeding the display limit are capped.}
\label{fig:tracking_result}
% \amin{you can move part of it to apendix?}\joe{Wenzhen said we can keep it here.}
\end{figure*}

We evaluate the 6DoF tracking error of all methods using the Mean Absolute Error (MAE) metric. Table \ref{tab:tracking_result} reports the average pose MAE of each method across all episodes, excluding those collected with the Ball and the Cylinder. These two objects are perfectly symmetrical and textureless, resulting in failure across all methods. Overall, GS-Online and GS-Offline achieve the best accuracy, and their close performance indicates that running in real time does not compromise GelSLAM's tracking quality. Our approach (GS-Online) reduces the rotation error by $46\%$ and the translation error by $17.5\%$ compared to NormalFlow (NF) by correcting drift through loop closures, while NF performs better than other frame-to-frame baselines (ICP, FilterReg, and FPFH+RI). The full SLAM baseline, Tac2Structure, performs poorly because our dataset includes many low-textured objects, which severely degrade the performance of its tracking module (FPFH+RI). Also, its loop detection method is not robust to outlier loops.

%Our method also outperforms the full SLAM baseline, Tac2Structure, demonstrating the robustness of our loop detection strategy.

\revised{We organize our ablation study in Table \ref{tab:tracking_result} around three key engineering decisions, also highlighting the impact of differential representations in our SLAM components.}

\noindent \textbf{Keyframe selection method:}
Without loop closure, using our keyframe selection method (NF-KF) achieves better results than the baseline without keyframes (NF). Introducing keyframes helps reduce drift by avoiding dense frame-to-frame composition and also lowers the computation in the loop closure module. Compared to the earlier keyframe selection method (NF-OrigKF) \cite{huang24}, \revised{our approach (NF-KF) uses differential-representation-based metrics (CCS and SCR)} to achieve similar performance while running twice as fast, requiring only a single NormalFlow computation per frame instead of two.

\noindent \textbf{Loop detection method:}
Incorporating loop closure using SIFT matching alone (GS-OnlySIFT) does not improve performance relative to our keyframe-based tracker (NF-KF) and can even degrade it due to outlier loops. Using NormalFlow alone for loop detection (GS-OnlyNF) achieves better results because our NormalFlow failure detection mechanism, \revised{based on the proposed CCS and SCR metrics,} can reject incorrect loop closures. However, it is significantly slower, as it must compare all pairs without SIFT-based pre-selection, which is about 40 times faster than NormalFlow. Moreover, NormalFlow without a good initialization can easily get stuck in local minima and falsely reject valid loops. The best results come from the two-stage pipeline (GS-Offline), where SIFT matching provides a reliable initialization, enabling NormalFlow to detect loops it would otherwise miss.

\noindent \textbf{SIFT feature source:}
We also study how the source of SIFT features affects robustness. \revised{Loop detection using SIFT extracted from the raw GelSight image (GS-ImageSIFT) is less reliable due to the image’s sensitivity to spatial transformations, while SIFT from a differential representation (GS-Offline, using the curvature map) improves matching reliability, reducing rotational tracking error by $34\%$.} Note that, unlike GS-OnlySIFT, GS-ImageSIFT includes the NormalFlow refinement stage for loop detection.

Beyond average performance across all objects, we also compare tracking results for each object on four representative methods: ICP, NormalFlow (NF), Tac2Structure, and our approach (GS-Online). The results are shown in Fig. \ref{fig:tracking_result}. All methods perform poorly on the Ball due to its perfectly smooth and symmetrical geometry. On the remaining objects, GS-Online achieves the best performance and shows clear improvement over NormalFlow on most objects. However, no improvement is observed on low-textured objects such as Block and Can, where too few valid loops can be detected. Tac2Structure performs well only on the Small Seed and Large Seed, which are highly textured, but frequently includes false loops on all other objects, resulting in poor performance. In contrast, GS-Online robustly detects valid loops when texture is present and avoids false detections when texture is limited. We also note that Tac2Structure operates offline.

\begin{comment}
\revised{\subsection{Effect of Constant Velocity Prior}}\label{sec:constant_velocity_prior}
\revised{Our work focuses on using tactile input alone to demonstrate that accurate long-horizon tracking is possible with minimal assumptoins, without relying on external sensors or motion priors. Here, we show that GelSLAM remains compatible with such priors and can benefit from them when available. We evaluate GS-Offline with a constant velocity prior following \cite{sodhi22}: [Equation Here] This results in an X\% reduction in rotation error and a Y\% reduction in translation error.}
\mytodo{Do the experiment and study write the equation down.}
\end{comment}

\revised{\subsection{Case Studies on Long-Horizon Tracking}}

\begin{figure}[t]
\centering
\includegraphics[width=0.98\linewidth]{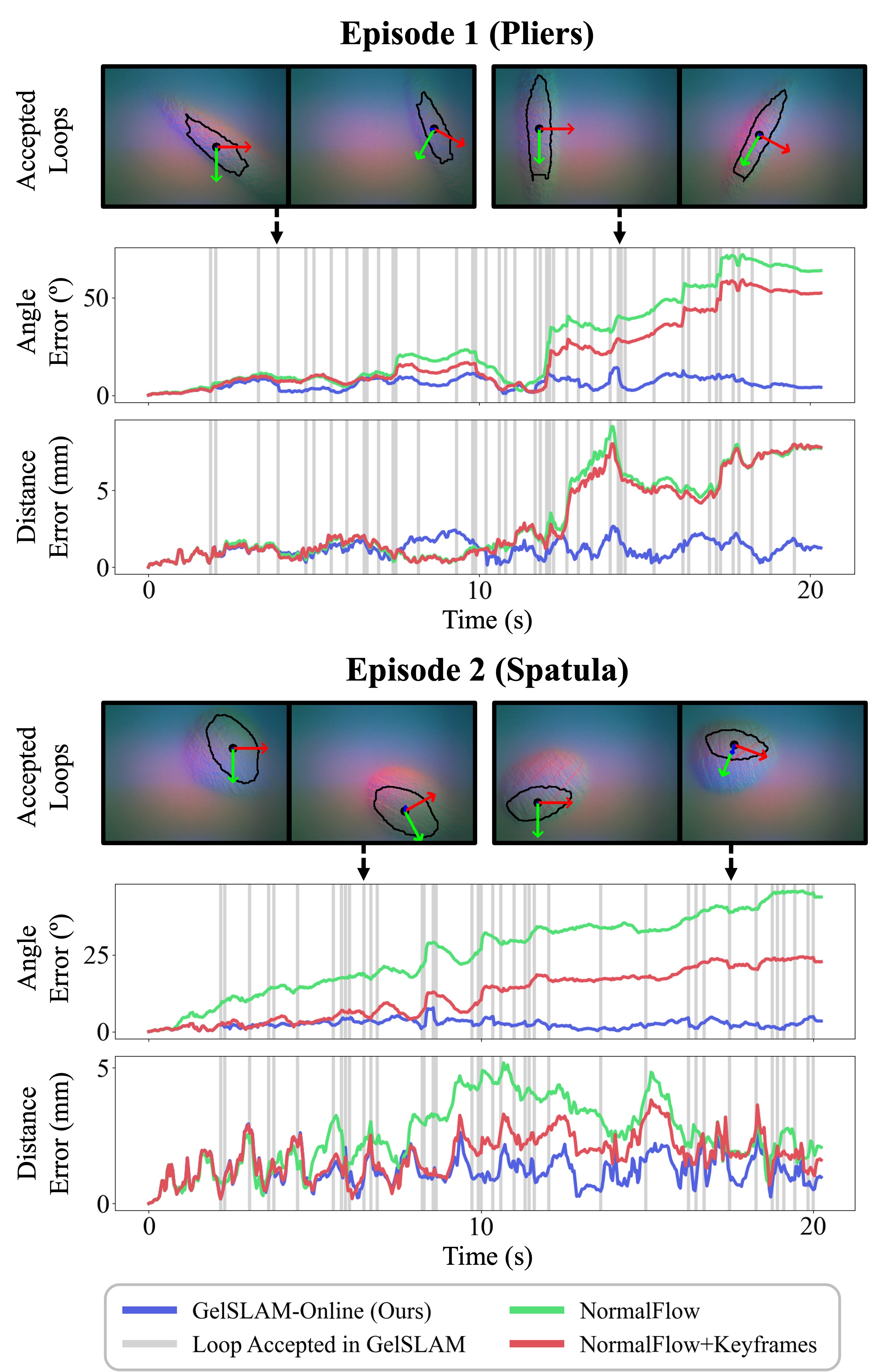}
\caption{Tracking error comparison between GelSLAM-Online, NormalFlow (NF), and NormalFlow+Keyframes (NF-KF) across two episodes. Gray vertical lines indicate when GelSLAM-Online detects and accepts loop closures. The top row of each episode shows examples of accepted loops as keyframe pairs with aligned coordinate systems. Well-matched textures in the shared contact region (black areas) confirm that the loops are correctly identified. Despite the low texture of the objects, GelSLAM-Online successfully detects many loops and effectively corrects drift during tracking.}
\label{fig:tracking_error_comparison}
\end{figure}

\begin{figure}[t]
\centering
\includegraphics[width=0.98\linewidth]{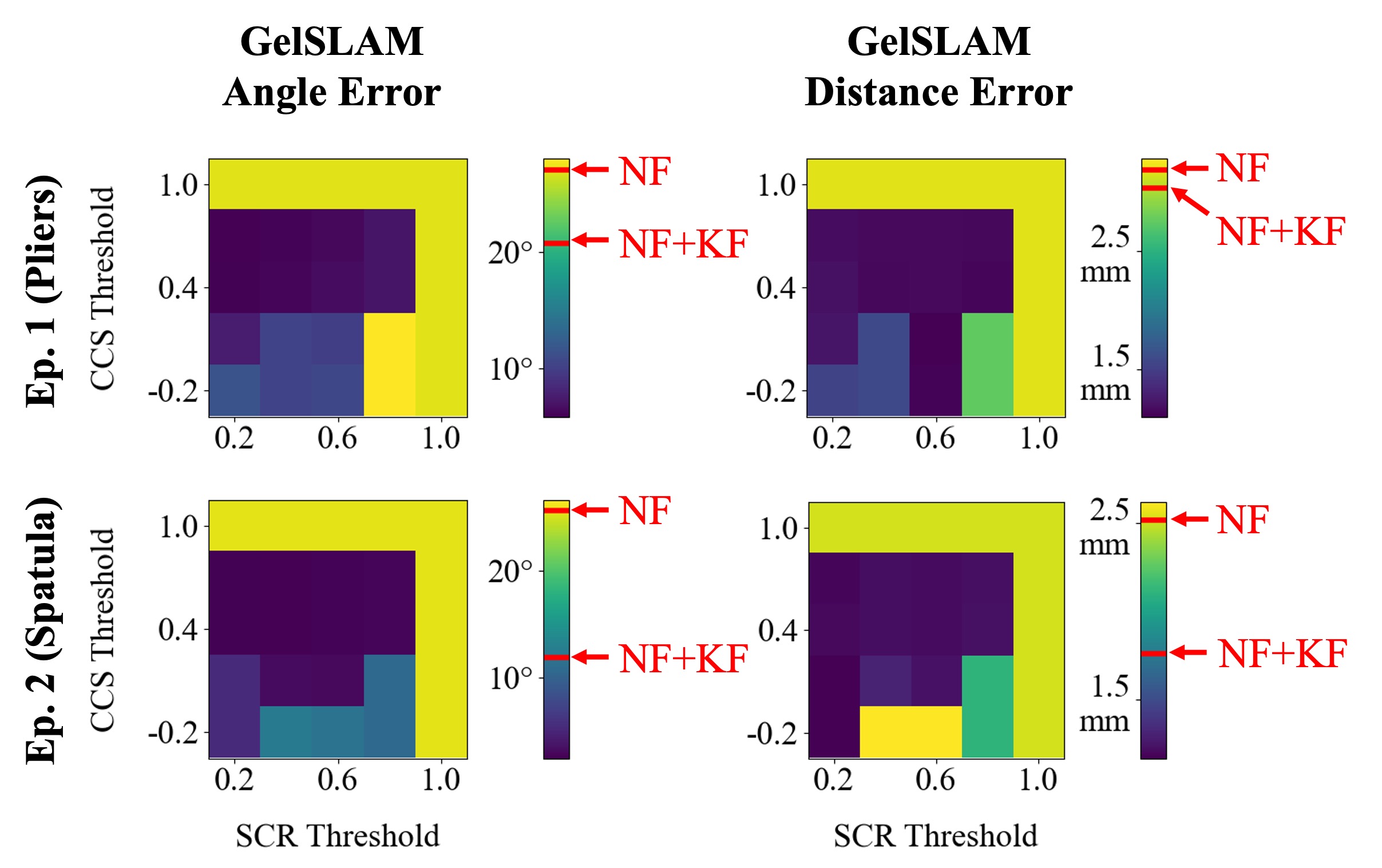}
\caption{\revised{Tracking error of our approach averaged over time across CCS and SCR thresholds. Tracking error is high at extreme thresholds but remains low across a wide range, with values much lower than the baselines indicated on the colorbar.}}
\label{fig:ablate_thresholds}
\end{figure}

\revised{We study two representative episodes (one on Pliers and one on Spatula) to analyze how the tracking error of our approach evolves over time, evaluate its sensitivity to the CCS and SCR thresholds, and examine its relocalization capability. Fig. \ref{fig:tracking_error_comparison} compares tracking error over time for NormalFlow (NF), NormalFlow with keyframes (NF-KF), and our method. Despite limited surface texture, our approach reliably identifies many loops and corrects accumulated drift, achieving consistently lower tracking error than the baselines.}

\revised{Fig. \ref{fig:ablate_thresholds} illustrates the sensitivity of our approach to the CCS threshold $(\in [-1,1])$ and SCR threshold $(\in [0,1])$. Setting either threshold to 1 makes every frame a keyframe and rejects all loop closures, causing GelSLAM to degenerate to NormalFlow. Across a wide range of CCS and SCR thresholds, the tracking error remains low, showing that performance is insensitive to the exact threshold choice. When the CCS threshold is too low (below 0), accuracy degrades due to accepting false loop closures; in practice, objects with less surface texture require a higher CCS threshold to avoid this issue. For long-horizon tracking, we find that a CCS threshold of 0.85 and an SCR threshold of 0.3 perform well across most everyday objects.} 

\begin{figure}[t]
\centering
\includegraphics[width=0.98\linewidth]{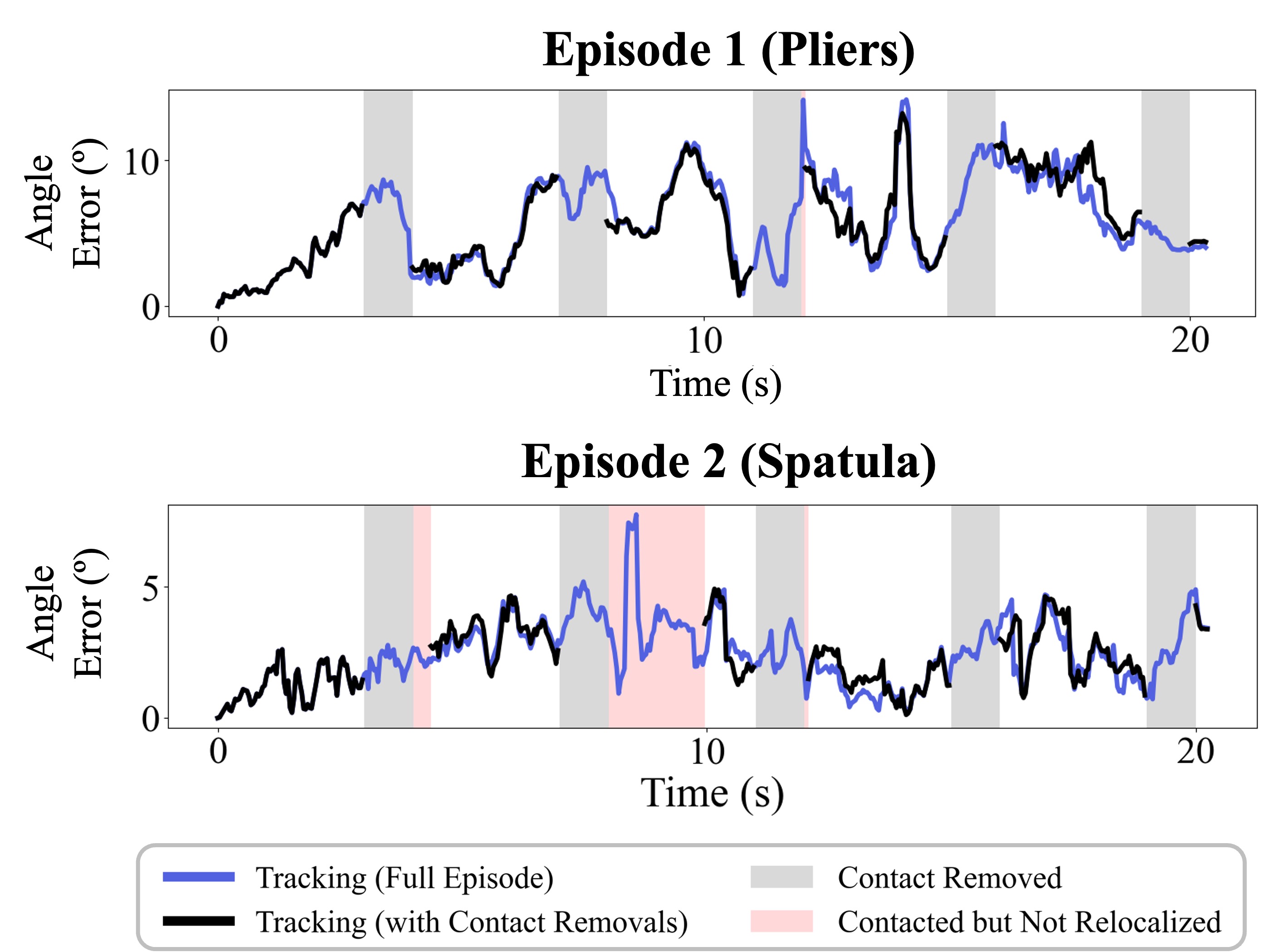}
\caption{\revised{Tracking error comparison of our approach with and without simulated contact loss. Our method can reliably relocalize after contact loss, achieving tracking performance nearly identical to full-episode results.}}
\label{fig:simulate_breaks}
\end{figure}

\revised{We evaluate the relocalization capability of our method by simulating periodic contact loss. Specifically, every 4 seconds (100 frames), we remove a 1-second segment (25 frames) from the sequence. Fig. \ref{fig:simulate_breaks} compares the tracking results of our approach obtained using the full episodes with those obtained under simulated contact removals. Once contact is re-established, our method reliably relocalizes and achieves tracking performance nearly identical to that obtained without contact removal, although relocalization may be delayed when contact resumes far from previously contacted regions.}

%\note{Figure 10: The comparison plot between ICP, NormalFlow, Tac2Structure and GS-Online (Ours) for each object.}

\section{Experiment: Object 3D Reconstruction}
In this section, we evaluate the qualitative and quantitative performance of GelSLAM on object-level 3D reconstruction. % across multiple objects.

\subsection{Data Collection}
\begin{figure}[htbp]
\centering
\includegraphics[width=0.7\linewidth]{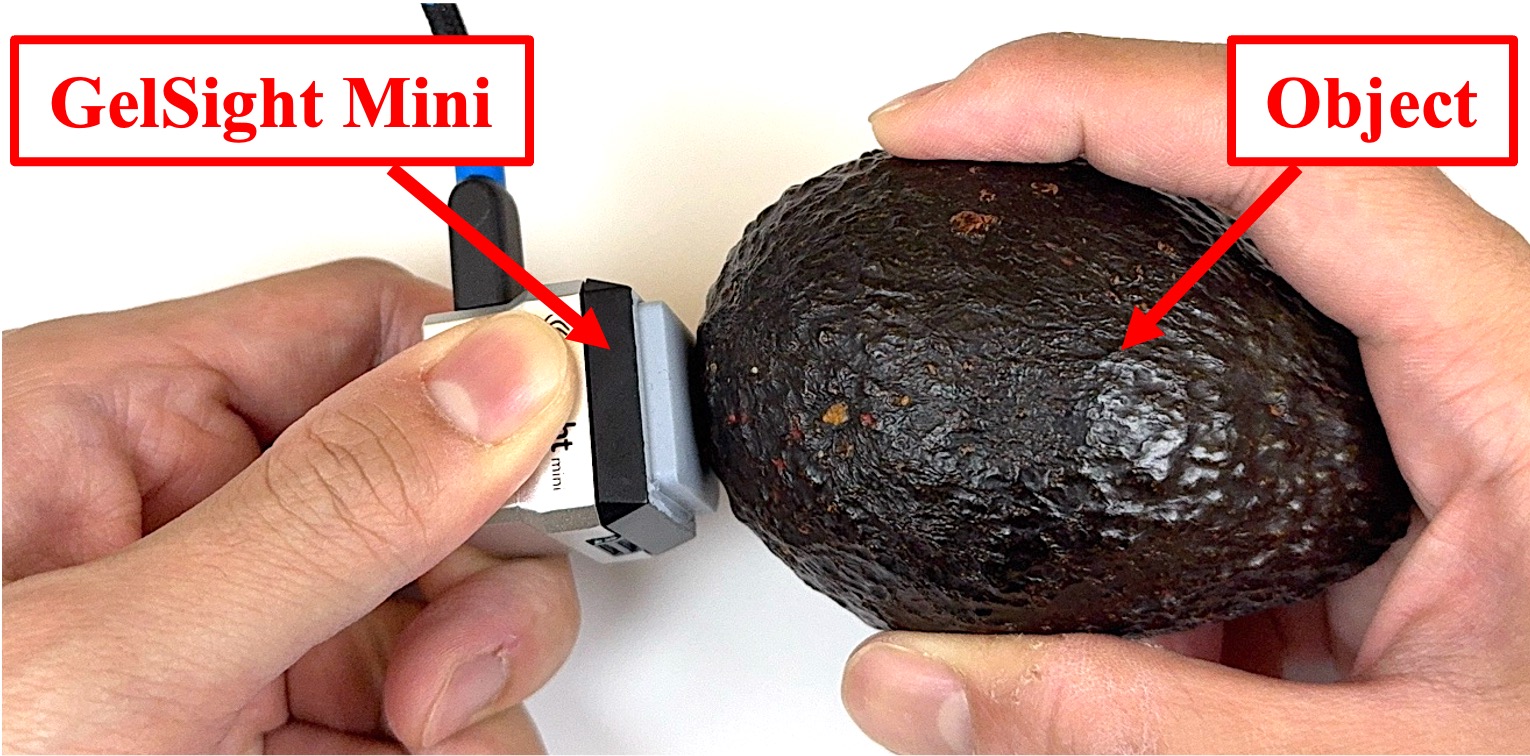}
\caption{Reconstruction data collection setup: both the GelSight sensor and the object are handheld, allowing free movement and contact breaks at any time.}
\label{fig:reconstruction_setup}
\end{figure}

\begin{figure*}[htbp]
\centering
\includegraphics[width=0.95\linewidth]{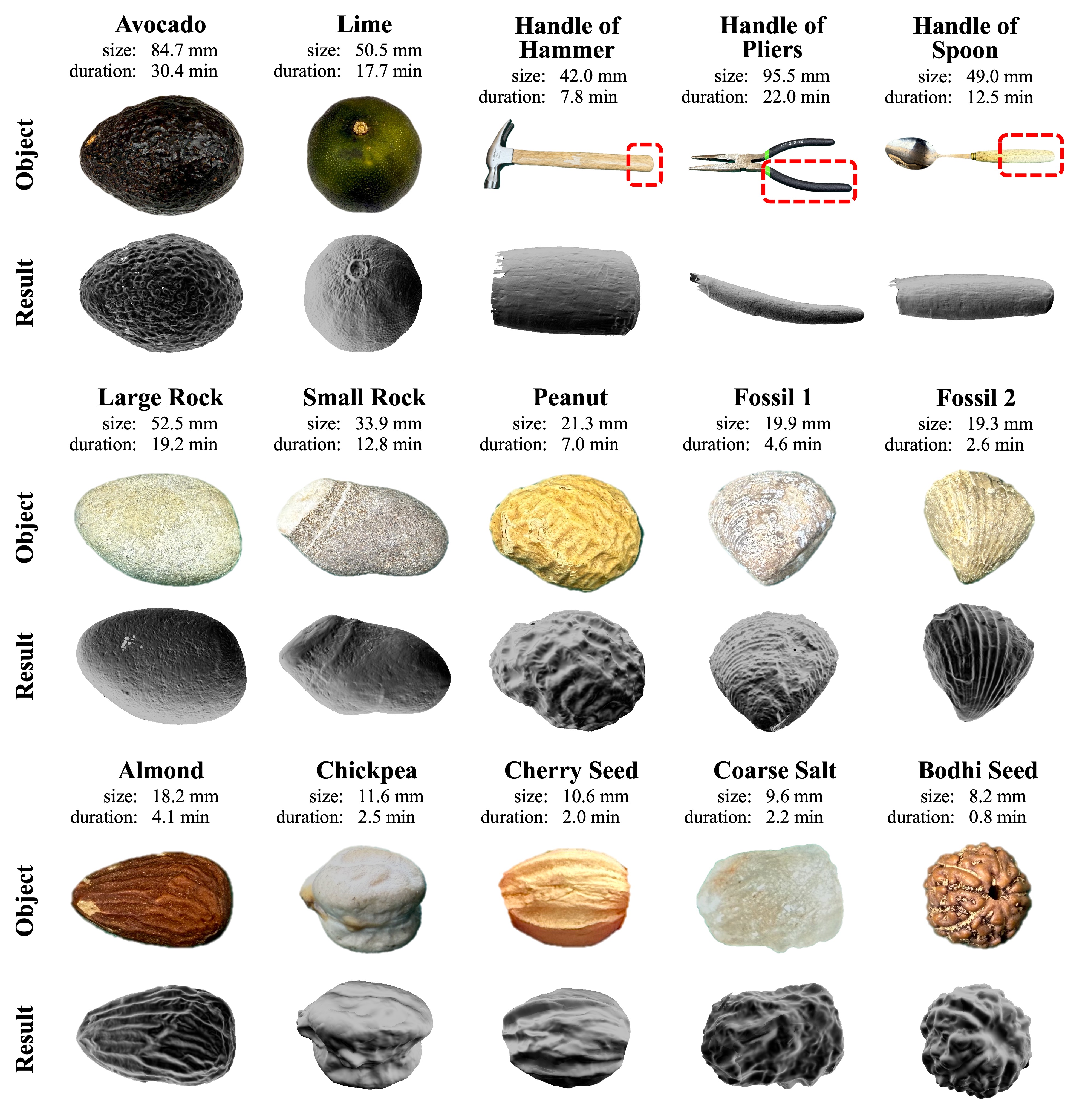}
\caption{GelSLAM reconstruction results for the $15$ objects, ordered roughly from largest to smallest. Object size and video duration are shown. GelSLAM successfully reconstructs detailed global 3D models using only local tactile patches.}
\label{fig:reconstruction_result}
\end{figure*}

We collect one GelSight video per object for qualitative analysis on a set of 15 real-world objects, including $3$ tool handles, $7$ common food items, $4$ rocks and fossils, and $1$ textured object. The smallest object is Seed (\(8\,mm \times 8\,mm \times 8\,mm\)), and the largest is Avocado (\(85\,mm \times 61\,mm \times 58\,mm\)). All objects have at least some surface texture, with the smoothest being the handle of a wooden spoon. Video durations range from $1$ to $30$ minutes. Fig. \ref{fig:reconstruction_result} shows each object with its size and video duration. For quantitative analysis, we use 5 object models with known ground-truth geometries. Each model is 3D-printed in two sizes using a Form 3+ printer, with a reported average dimensional deviation of up to 0.1\,mm, resulting in a total of 10 printed objects. The smallest and largest objects are a Seed and Lime with a diameter of about 9\,mm and 30\,mm, respectively. All data are collected in an ``in-the-wild" setup, where we manually hold both the object and the GelSight Mini sensor (Fig. \ref{fig:reconstruction_setup}) during scanning. Contact breaks and re-initializations are common during data collection, sometimes exceeding 100 occurrences per object. \revised{Scanning paths are guided by real-time reconstruction visualization, which the operator uses to adjust sensor movement during scanning. For very long scans, we segment data collection into consecutive 10-minute scans and fuse the resulting tactile videos into a single sequence for reconstruction.}

%\note{Figure 11: In-the-wild data collection setup for 3D reconstrutcion.}

\subsection{System Characteristics and Performance}
\begin{figure}[htbp]
\centering
\includegraphics[width=0.95\linewidth]{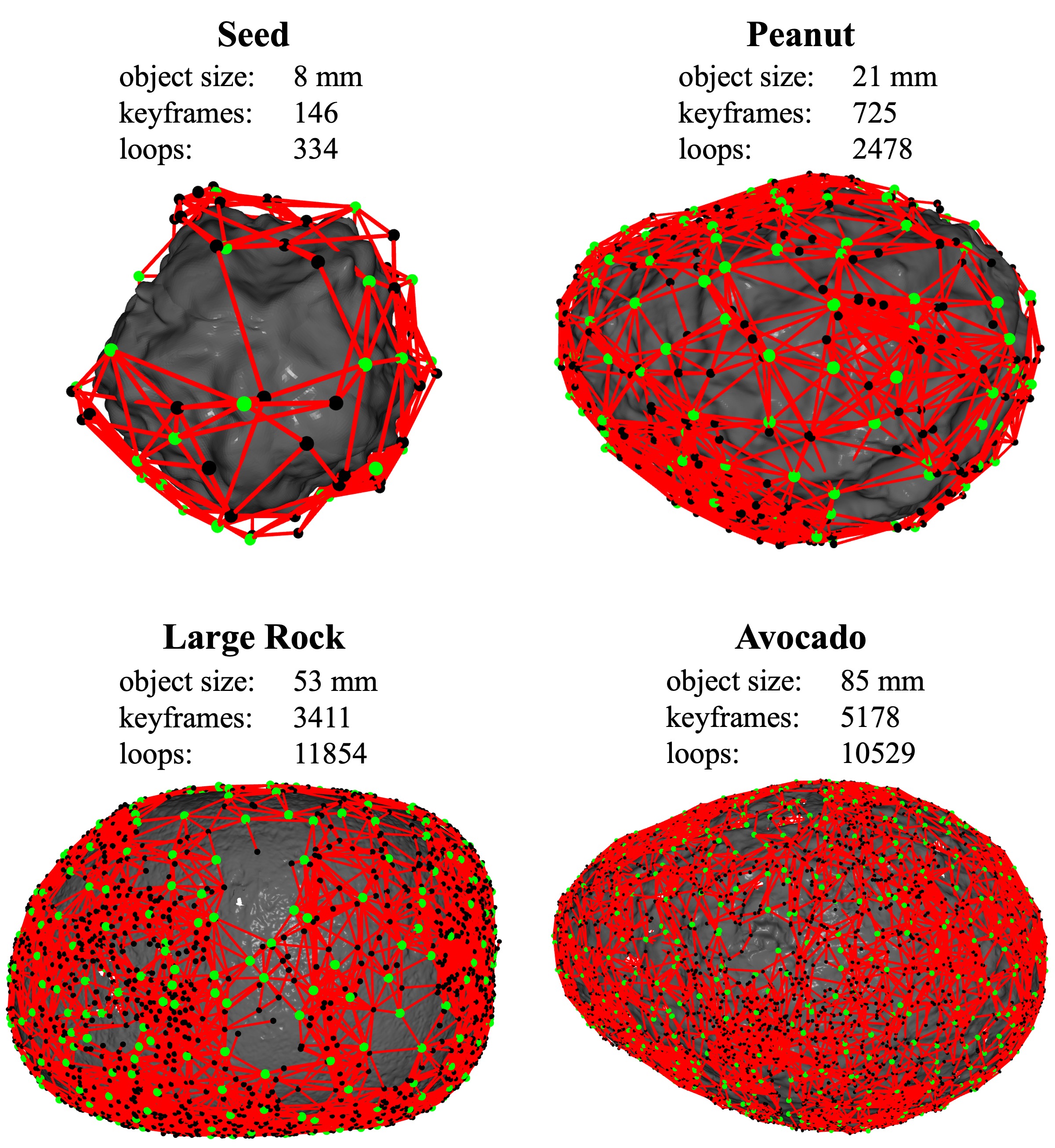}
\caption{Pose graphs generated by GelSLAM for four representative objects. Black nodes: keyframes; green nodes: coverage keyframes; red edges: pairwise pose constraints.}
\label{fig:pose_graphs}
\end{figure}

We run GelSLAM in offline mode on all object videos. \revised{We focus on offline reconstruction here, as offline shape reconstruction is sufficient for most reconstruction applications, and discuss online reconstruction in Section \ref{sec:online_reconstruction}.} Since the objects provide sufficient texture, we apply fixed and less conservative thresholds for NormalFlow failure detection: the estimate is rejected if $\text{CCS} < 0.7$ or $\text{SCR} < 0.3$ across all objects. Table \ref{tab:reconstruction_statistics} summarizes reconstruction statistics for four objects with varying video lengths, and Fig. \ref{fig:pose_graphs} shows their corresponding pose graphs.

\begin{figure}[htbp]
\centering
\includegraphics[width=0.95\linewidth]{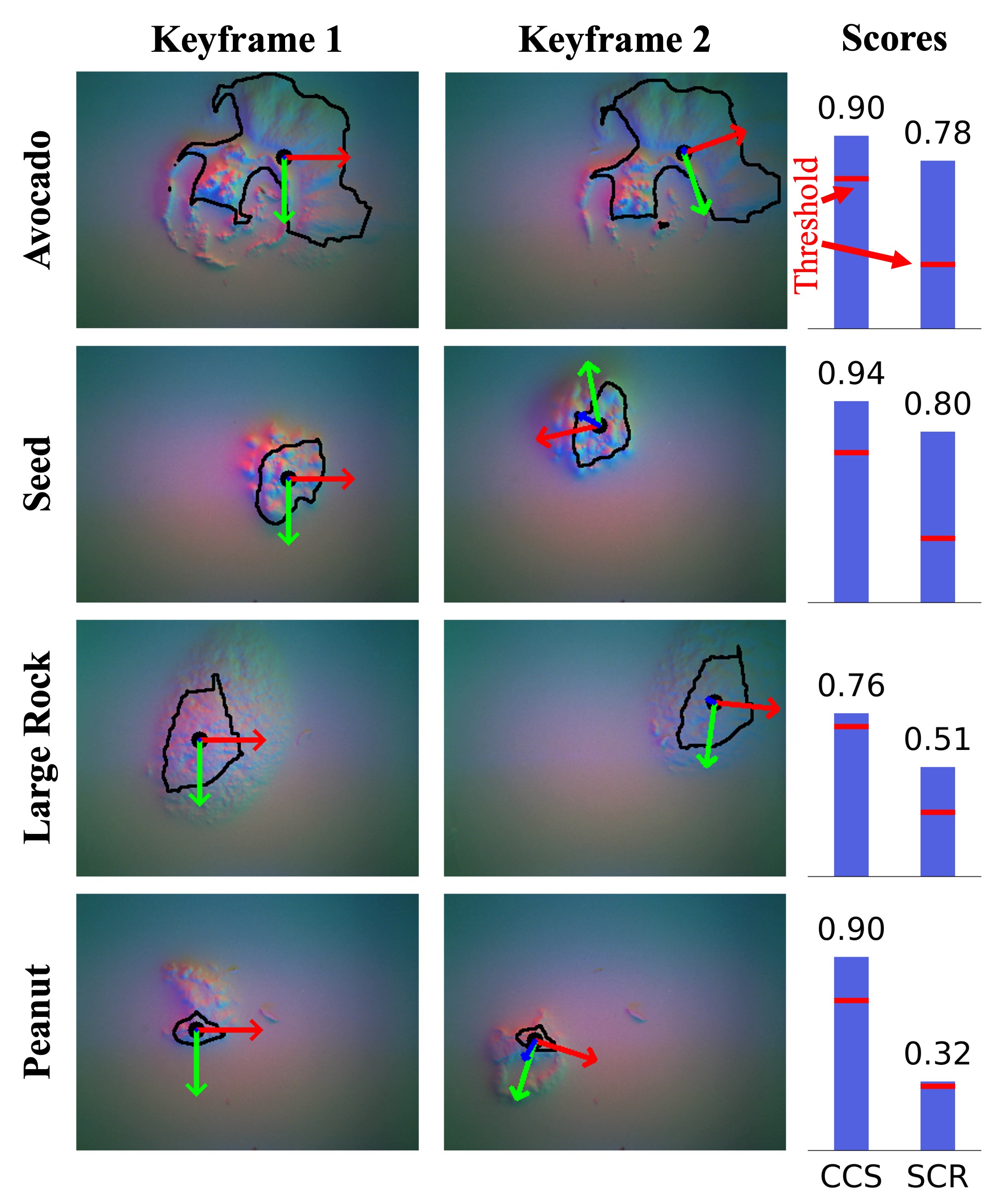}
\caption{Example of accepted loops. \textbf{[Left and middle]} Keyframe pair with aligned coordinate systems. Textures are well aligned within the black regions, which indicate shared contact areas. \textbf{[Right]} Curvature Cosine Similarity (CCS) and Shared Curvature Ratio (SCR) scores for the loop. Loop is accepted as both scores exceed the threshold (red line). Our method reliably detects loops despite large rotation (second row), low texture (third row), and little overlap (fourth row). \joe{Put the conclusion in the caption, put the full name of the metric in the caption. Put legend for the threshold.}\joe{Done?}}
\label{fig:reconstruction_loop_detection}
\end{figure}

\begin{figure*}[htbp]
\centering
\includegraphics[width=0.93\linewidth]{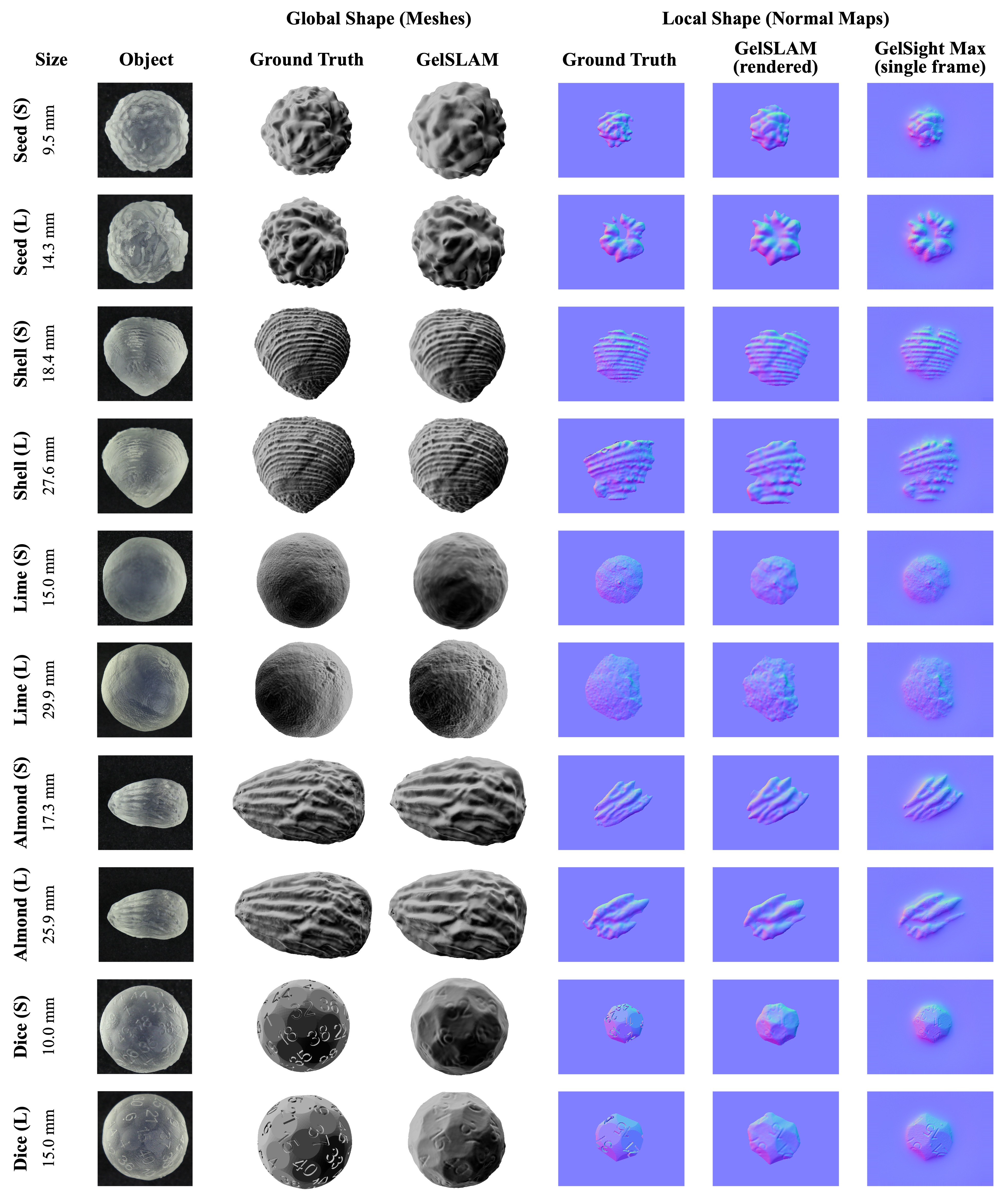}
\caption{Comparison between GelSLAM-reconstructed shapes and ground truth CAD models of 3D-printed objects. Five models were each printed in two sizes, labeled small (S) and large (L). For each object, from left to right: object size, external view (resized to fit), ground truth mesh (CAD model), GelSLAM mesh, ground truth texture (normal map from CAD), GelSLAM texture (normal map from reconstructed mesh), and GelSight Max texture (single-frame scan from the micron-level accuracy scanner). The GelSLAM-reconstructed mesh closely matches the ground truth CAD model in global geometry and recovers local textures with only a slight loss of sharp detail. Differences between ground truth and GelSight Max textures reflect imperfections introduced during the 3D printing process.}
\label{fig:reconstruction_quantitative_result}
\end{figure*}

\begin{table}[ht]
\renewcommand{\arraystretch}{1.2}
\centering
\begin{tabular}{|c|c|c|c|c|}  %<-- change here
\hline 
\textbf{Object Name} & \textbf{Avocado} & \makecell{\textbf{Large} \\ \textbf{Rock}} & \textbf{Peanut} & \textbf{Seed}  \\ 
\hline 
\hline 
Object size (mm) & 84.7 & 52.5 & 21.3 & 8.2   \\ 
\hline 
Video Length (min) & 30.37 & 19.18 & 6.97 & 0.75   \\ 
\hline 
Frames & 45557 & 28779 & 10470 & 1211   \\ 
\hline 
Tracking Sessions & 442 & 324 & 57 & 8   \\ 
\hline 
Keyframes & 5178 & 3411 & 725 & 146   \\ 
\hline 
Coverage Keyframes & 1297 & 511 & 202 & 40   \\ 
\hline 
Loop Candidates & 12851 & 23041 & 2957 & 397   \\ 
\hline 
Loops & 10529 & 11854 & 2478 & 334   \\ 
\hline 
\makecell{GelSLAM Offline \\ Runtime (min)} & 40.92 & 29.82 & 4.78 & 0.6   \\ 
\hline 
\end{tabular}
\caption{GelSLAM statistics for object-level 3D reconstruction on four objects of varying sizes.}
\label{tab:reconstruction_statistics}
\end{table}

Tactile reconstruction produces significantly more loop closures than typical indoor SLAM \cite{murartal15}. In indoor environments, loops typically form at specific topological features such as corridor or road intersections. In contrast, tactile keyframes are densely distributed across the object surface, enabling loop closures to occur wherever the sensor revisits a region. This frequent loop formation is essential for tactile reconstruction due to the inherently local nature of touch. Moreover, the high number of tracking sessions in tactile scanning reflects frequent re-localization, further increasing reliance on robust loop detection. These challenges require robust outlier rejection using only local information, which GelSLAM demonstrates to be effective. In Table \ref{tab:reconstruction_statistics}, none of the detected loops were false positives, as even a single outlier can lead to catastrophic behavior. Fig. \ref{fig:reconstruction_loop_detection} shows four examples of the detected loops. Even under challenging conditions such as large object rotation (second row), low surface texture (third row), and minimal overlap (fourth row), our loop detection method reliably identifies valid loops.

%\revised{While reconstruction use cases only require GelSLAM to run offline, we run it online and see its performance. We further analyze when loop detection is skipped during online reconstruction and its impact on reconstruction quality. Fig. X plots the loop detection runtime together with the time interval before the next keyframe is introduced during a 30-minute scanning episode. Early in the sequence, loop detection consistently completes before the next keyframe arrives, whereas later in the episode, loops are skipped intermittently as the system scales. Fig. Y shows the resulting online reconstruction, which remains nearly identical to the offline reconstruction despite skipped loops, even for larger objects. Finally, we note that reconstruction can always be performed offline; this analysis is intended to characterize the runtime behavior and performance of online GelSLAM as the number of frames increases.}

%\note{Figure 12: Two-columned figure, two rows, first row: pose graph, second row: session graph. Four columns, each with an object. Maybe we can also present a zoom-in to see the detailed reconstruction.}

%\note{Figure 13: Five columns, ref image, ref curvature map, target image, target curvature map, curvature map difference Li'-Lj. Showing four loop closure results.}

\subsection{Qualitative Results}

Fig. \ref{fig:reconstruction_result} shows tactile-based object-level 3D reconstructions for the $15$ real-world objects. All loops detected by GelSLAM are correct, indicating its robustness. The results demonstrate that GelSLAM correctly relates GelSight images spatially and globally reconstructs objects across a wide range of textures and sizes, even on low-texture objects such as the wooden spoon handle. The reconstructed geometry also exhibits remarkable detail.

\begin{figure}[htbp]
\centering
\includegraphics[width=0.95\linewidth]{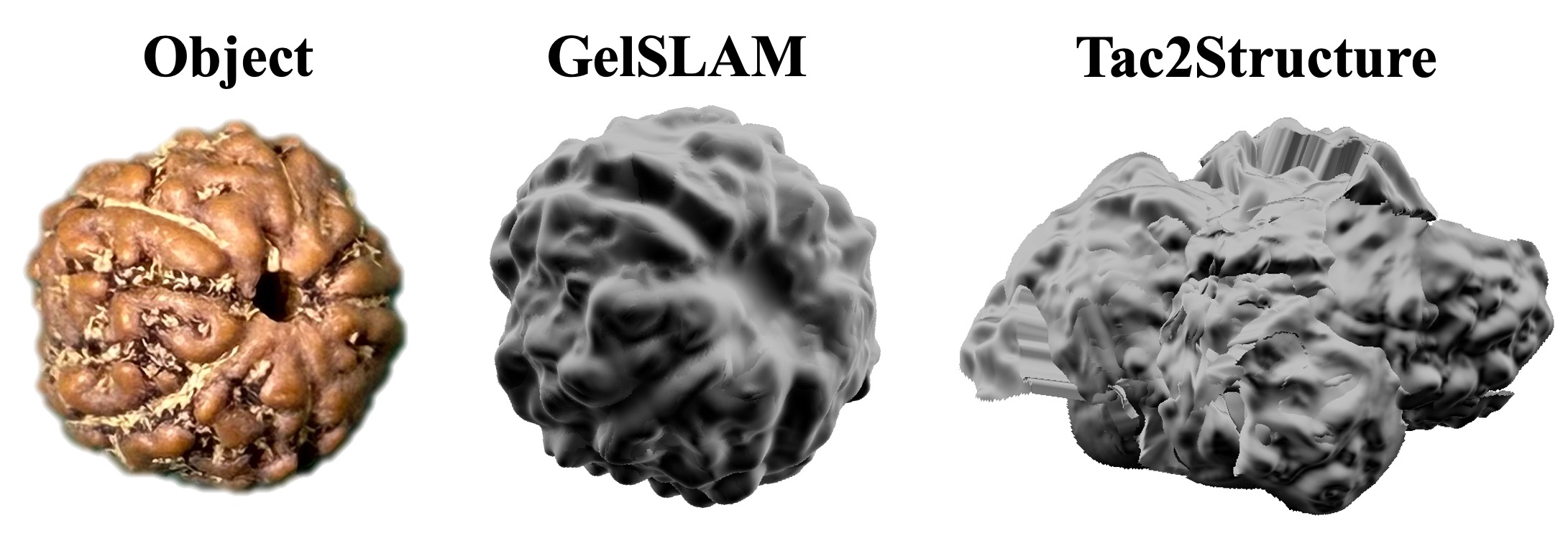}
\caption{Reconstruction result of GelSLAM and Tac2Structure on the Seed. Tac2Structure includes many false positive loops, resulting in a failed reconstruction.}
\label{fig:reconstruction_tac2structure}
\end{figure}

We compare GelSLAM with Tac2Structure \cite{lu23} in Fig. \ref{fig:reconstruction_tac2structure} using the Seed, the easiest object to reconstruct in our dataset due to its rich texture and fewer frames. Despite this, Tac2Structure fails to reconstruct the object. Its CNN-based loop detection method is prone to including false loops, especially when detecting loops across tracking sessions, leading to reconstruction failure. In contrast, GelSLAM robustly identifies correct loops and achieves accurate reconstruction results.

%\note{Figure 14: A two-columned figure, with all 17 object with names, object image, a front view and a back view of the reconstructed mesh, and a scale. We can put them in four scales, the top row with smallest scale (equal across the row) and the bottom with largest scale (equal across the row)}
%\note{Figure 15: Robust solver on Round Bead v.s. non-robust solver.}
%\note{Figure 16: Comparison with Tac2Structure on three objects.}
\begin{comment}
\begin{itemize}
    \item For each object, show the object image, object size, reconstructed front side, back side, video length. Explicitly mention that bead is solved with robust solver.
    \item Show reconstruction result of Tac2Structure on 3 objects and compare to that reconstructed by our algorithm. \joe{Should we compare to vision-based methods? It's hard because we are not using the same dataset. We need to collect another dataset specifically for vision, which is hard.}
    \item Comment on our qualitative results. 
\end{itemize}
\end{comment}

\subsection{Quantitative Results}
Fig. \ref{fig:reconstruction_quantitative_result} shows the 3D reconstruction results for ten 3D-printed objects. For global shape, we show that the meshes reconstructed by GelSLAM align well with the corresponding ground truth CAD models. For local shape, we compare normal maps, which better capture fine surface details than meshes. Specifically, we show the ground truth texture (normal map rendered from the CAD model), the GelSLAM texture (normal map rendered from the reconstructed mesh), and a single scan from the GelSight Max, a commercial scanner with micron-level accuracy. The GelSLAM texture closely matches the ground truth texture, though with a slight loss of fine detail. This is primarily due to the limited resolution selected for GelSight Mini as well as minor pose estimation errors that introduce slight misalignment between tactile patches, leading to averaging effects that smooth out fine details. The ground truth CAD texture also slightly differs from the GelSight Max scans due to imperfections in the 3D printing process.

Table \ref{tab:qualitative_reconstruction_results} presents quantitative comparisons between the GelSLAM reconstructions and the CAD models, evaluating both global and local shape similarity. For global shape, we report the Chamfer Distance (CD) between the reconstructed and CAD meshes. On average, the CD is $0.6$\,mm across all objects, whose average size is $18.3$\,mm. For local geometry, we evaluate surface texture similarity by comparing normal maps generated from both meshes. Specifically, we simulate GelSight Mini tactile observations at 100 uniformly distributed contact locations over the object surface, with touch directions aligned with the surface normal, and represent each observation in the form of a normal map to capture the local geometry. We then compute the Normal Cosine Distance (NCD), defined as the average per-pixel cosine distance between corresponding normal maps. A perfect match yields an NCD of $1.0$. Across all objects, the average NCD is $0.962$, indicating that the reconstructed textures closely match the CAD models.

\begin{table}[ht]
\renewcommand{\arraystretch}{1.2}
\centering
\begin{tabular}{|@{\hskip 4pt}c@{\hskip 4pt}|@{\hskip 4pt}c@{\hskip 4pt}|
                @{\hskip 4pt}c@{\hskip 4pt}|@{\hskip 4pt}c@{\hskip 4pt}|
                @{\hskip 4pt}c@{\hskip 4pt}|@{\hskip 4pt}c@{\hskip 4pt}|}
\hline 
& \textbf{Dice (S)} & \textbf{Shell (S)} & \textbf{Almond (S)} & \textbf{Seed (S)} & \textbf{Lime (S)}  \\ 
\hline 
size (mm) & 10.0 & 18.4 & 17.3 & 9.5 &  15.0 \\ 
\hline 
CD (mm) & 0.44 & 0.76 & 0.60 & 0.63 & 0.26  \\ 
\hline 
NCD & 0.955 & 0.944 & 0.966 & 0.954 & 0.985  \\ 
\hline 
\hline 
& \textbf{Dice (L)} & \textbf{Shell (L)} & \textbf{Almond (L)} & \textbf{Seed (L)} & \textbf{Lime (L)}  \\ 
\hline 
size (mm) & 15.0 & 27.6 & 25.9 & 14.3 & 29.9  \\ 
\hline 
CD (mm) & 0.41 & 0.91 & 0.85 & 0.63 & 0.50  \\ 
\hline 
NCD & 0.947 & 0.955 & 0.962 & 0.967 & 0.986  \\ 
\hline 
\end{tabular}
\caption{Chamfer Distance (CD) and Normal Cosine Distance (NCD) between the CAD model and GelSLAM reconstructed meshes for all 3D printed objects. }
\label{tab:qualitative_reconstruction_results}
\end{table}

\subsection{Large Object Reconstruction using GelBelt}
In this section, we show that GelSLAM can be applied to reconstruct larger objects and is compatible with vision-based tactile sensors other than GelSight Mini. We use the GelBelt sensor \cite{mirzaee25}, which features a sensing belt that enables continuous scanning of larger surfaces. The sensing belt rolls over two wheels, slides under the acrylic layer in front of the camera, while sticking to the in-contact surface. GelBelt’s sensing area is $60\,\text{mm} \times 40\,\text{mm}$, and we resize its resolution to $216 \times 288$ pixels for processing.

\begin{figure}[htbp]
\centering
\includegraphics[width=0.95\linewidth]{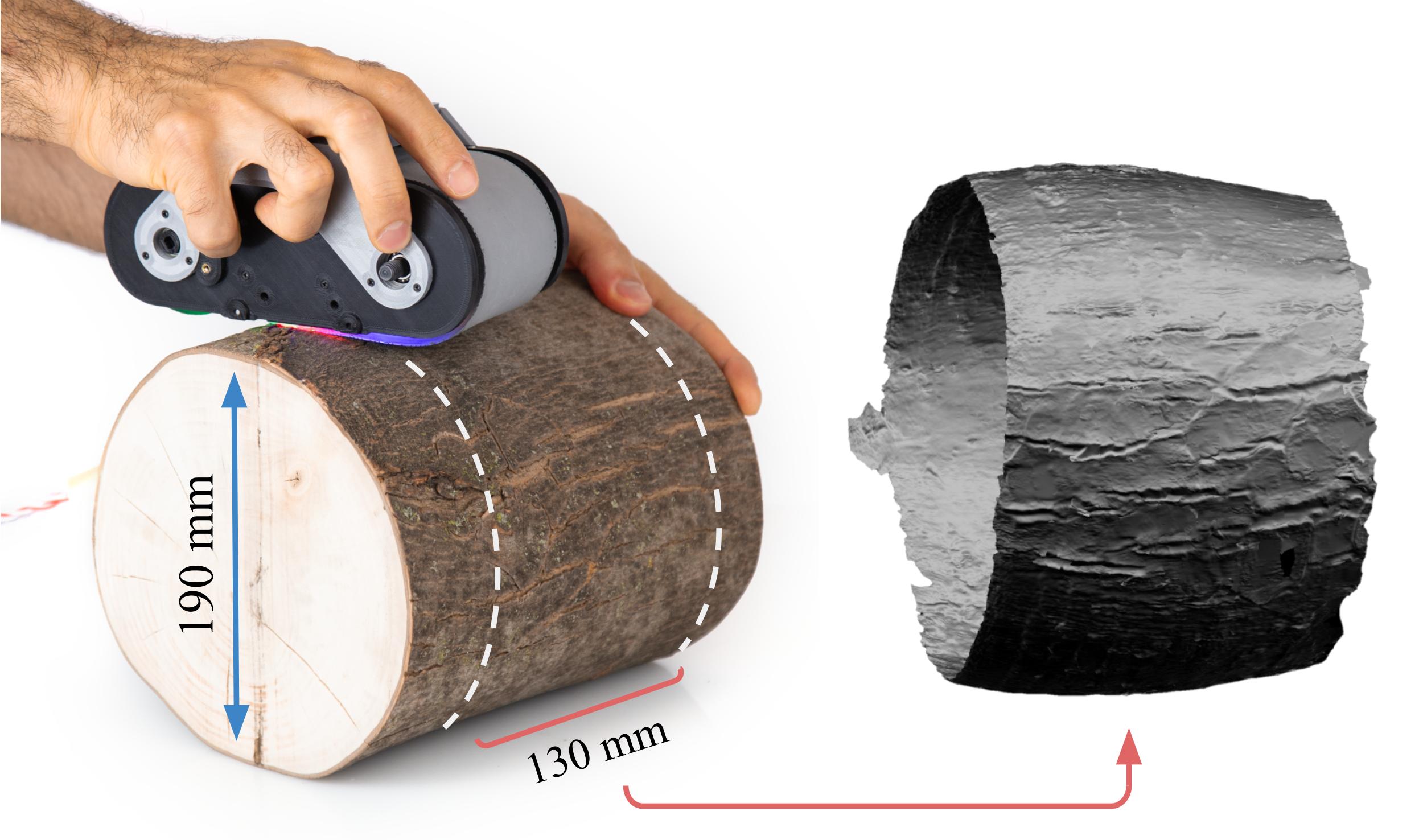}
\caption{Continuous scanning and reconstruction using the GelBelt sensor. The sensor is rolled over the tree trunk multiple times in both radial and vertical directions to collect data for reconstruction. The resulting mesh closely matches the real surface texture and overall shape of the trunk.}
\label{fig:gelbelt}
\end{figure}

We qualitatively evaluate the reconstruction of a tree trunk with a diameter of approximately 190\,mm by scanning around 130\,mm of its length using GelBelt. The sensor was rolled along the trunk surface, moving both vertically and radially, and completed nine full rounds around the trunk. Fig. \ref{fig:gelbelt} shows the resulting reconstruction. The reconstructed surface mesh closely matches the real texture of the trunk, including visible lines and cracks. \revised{We observe slight shrinkage near the two ends of the reconstructed cylindrical mesh. This is likely because the middle regions are constrained by pose graph edges from multiple directions, while the ends are constrained mainly by one-sided pose graph connections, making them more susceptible to global distortion.}

\section{Discussion}

\revised{Our experiments demonstrate that GelSLAM achieves real-time, robust long-horizon tracking, together with high-fidelity reconstruction, across a wide range of objects. At the same time, we identify several challenging conditions that motivate future research directions.}

\begin{figure}[htbp]
\centering
\includegraphics[width=0.7\linewidth]{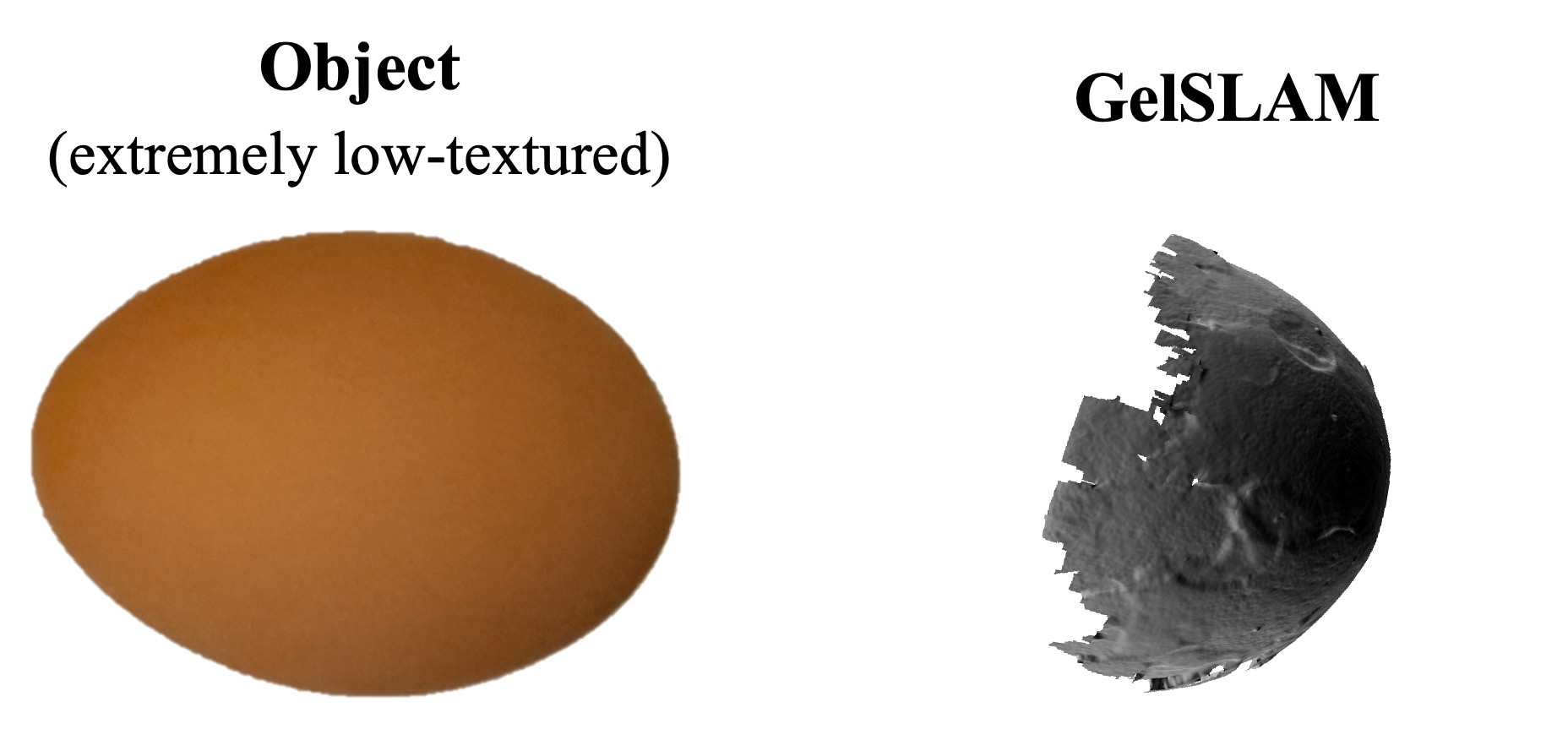}
\caption{\revised{Challenging scenario. For extremely low-textured objects, GelSLAM may fail to detect loop closures, resulting in fragmented and locally discontinuous reconstructions.}}
\label{fig:low_texture_reconstruction}
\end{figure}

\revised{\subsection{Challenging Conditions and Partial Solutions}}
\revised{For objects with extremely low surface texture, GelSLAM can struggle with loop detection, leading to pose drift, relocalization failure, and disconnected or locally discontinuous regions in the reconstructed mesh. Fig. \ref{fig:low_texture_reconstruction} shows the GelSLAM reconstruction of an egg, a representative near-textureless object. Only the largest connected component of the reconstructed mesh is shown. Although the entire surface is scanned, failure to relocalize after contact loss leads to disconnected mesh segments. In addition, the reconstructed mesh exhibits local discontinuities, which arise from uncorrected pose misalignment when regions are revisited but loop detection is missed. Despite these issues, GelSLAM still achieves a partial reconstruction, with $11$ tracking sessions, $221$ keyframes, and $70$ detected loops forming the largest connected mesh, suggesting meaningful relocalization and loop detection capability despite minimal surface texture. In general, touch alone is unlikely to resolve this limitation for extremely low-textured objects. Incorporating complementary sensing modalities, such as vision, may provide the global context needed for improved loop detection.}

\begin{figure}[htbp]
\centering
\includegraphics[width=0.95\linewidth]{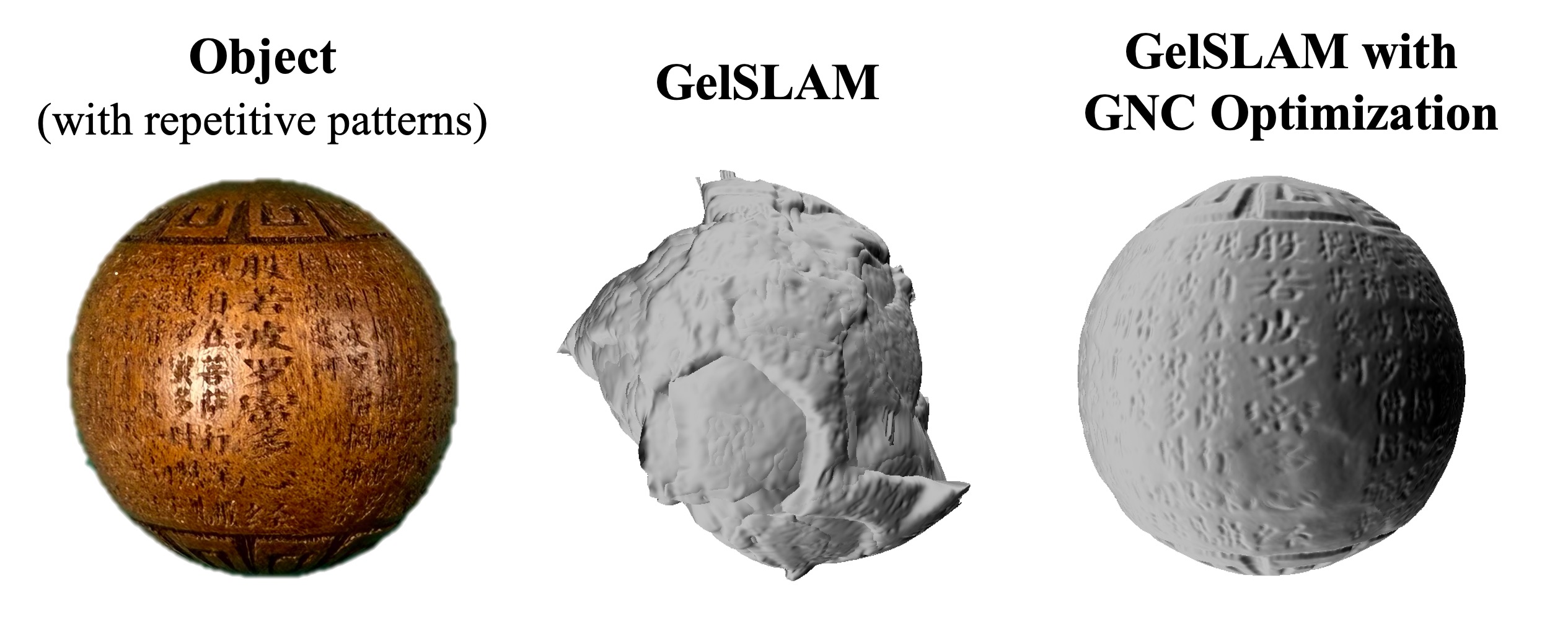}
\caption{\revised{Challenging scenario. For objects with repetitive patterns, GelSLAM may suffer from outlier loop closures. Applying GNC-based pose graph optimization filters these outliers and significantly improves reconstruction quality.}}
\label{fig:gnc_optimization}
\end{figure}

\revised{Another challenge arises for objects with highly repetitive surface patterns, where loop detection may incorrectly associate frames from different locations. Although such repetition is uncommon in GelSight sensing due to its high resolution, it can occur in industrially manufactured objects and lead to outlier loop closures that significantly degrade pose estimation and reconstruction. As an initial approach, we propose to apply a GNC-based pose graph optimization \cite{yang20} to reject outlier loops, which can improve reconstruction quality in these cases as shown in Fig. \ref{fig:gnc_optimization}. However, the increased computational cost and the fact that GNC can occasionally reject valid loops indicate that better solutions are still needed.}

\revised{\subsection{Online Reconstruction}} \label{sec:online_reconstruction}

\begin{figure}[htbp]
\centering
\includegraphics[width=0.95\linewidth]{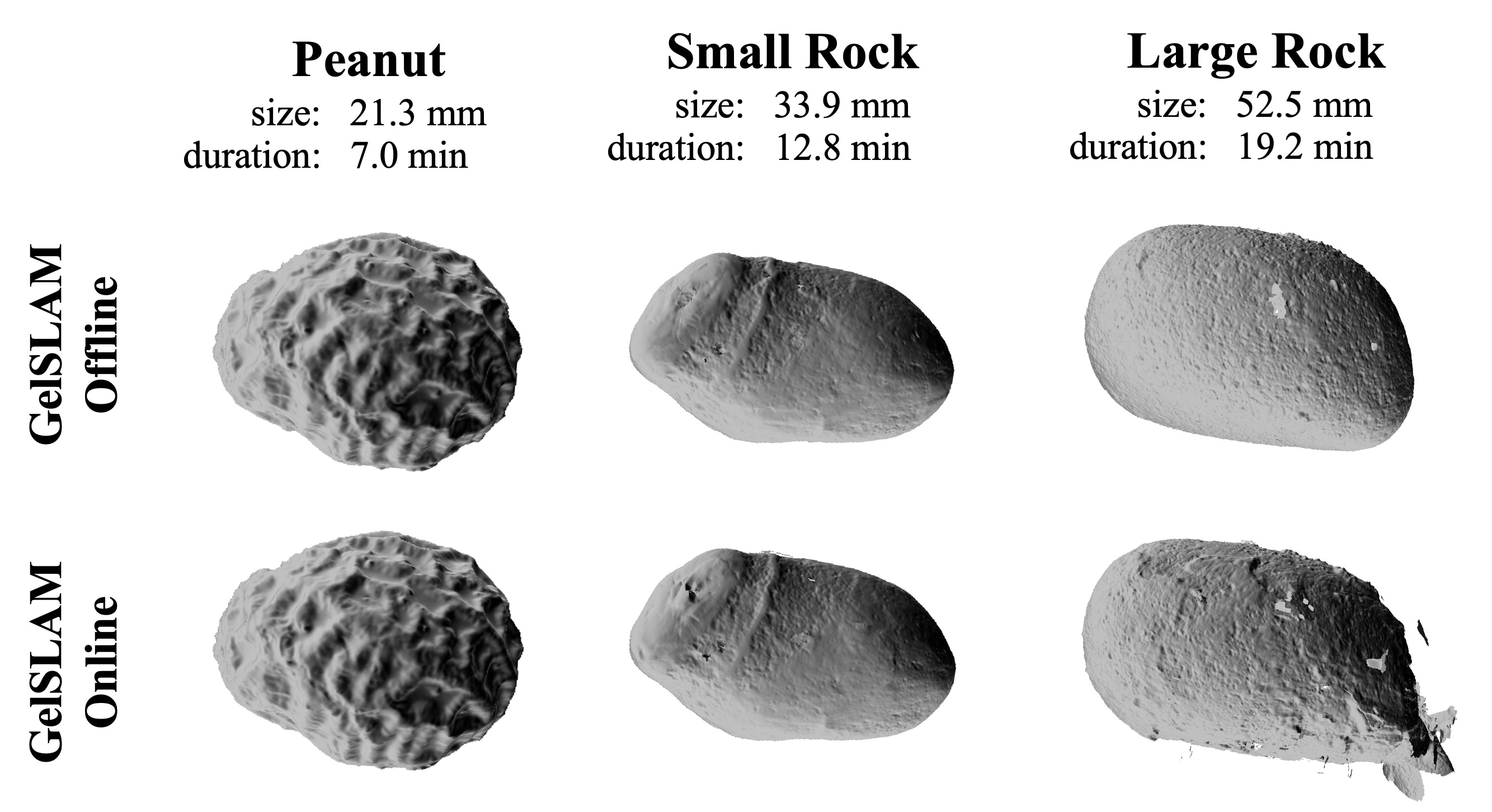}
\caption{\revised{Online vs. offline GelSLAM reconstruction. Online reconstruction closely matches offline results for 7- and 12-minute scans, but introduces artifacts for a longer 19-minute scan. In practice, reconstruction rarely needs to be real-time, and offline reconstruction meets most use cases.}}
\label{fig:online_reconstruction}
\end{figure}

\revised{GelSLAM can run in real-time for at least 10 minutes without noticeable performance drop. Since the primary real-time use case of GelSLAM is object pose tracking for manipulation, which typically lasts well under 10 minutes, this time span is more than sufficient in practice. For reconstruction, most applications do not require real-time mesh output and can therefore be performed offline. For the rare cases requiring online reconstruction, Fig. \ref{fig:online_reconstruction} shows that online results closely match offline reconstruction for 7-minute (peanut) and 12-minute (small rock) scans, but introduce artifacts for a 19-minute scan (large rock) due to loop skipping.}

%By nature, tactile sensing is highly local, resulting in much denser keyframes and loop closures than in typical SLAM systems so we think this can only.

\revised{\subsection{Future Directions}}

\revised{Several extensions could further improve the efficiency, robustness, and accuracy of GelSLAM. One promising direction is to reduce the need to touch the entire object surface by inferring untouched regions using prior knowledge or completing them with visual information. In addition, GelSLAM estimates contact masks via simple height thresholding, as in NormalFlow \cite{huang24} for its simplicity. One could adopt learning-based contact mask estimation methods \cite{suresh2022shapemap} for potentially better performance.}
% In this paper, a fixed height threshold is used across all experiments, except for reconstructing Lime and Avocado.  Applying this threshold to these larger objects has little effect on GelSLAM, but can introduce holes in the tactile meshes and, consequently, in the reconstructed mesh. In addition, GelSLAM estimates contact masks using simple height thresholding, adopting the approach used in NormalFlow \cite{huang24} for its simplicity. For large objects, this approach can misclassify true contact regions as non-contact, introducing holes in the reconstructed mesh. Simply replacing it with learning-based contact mask estimation methods \cite{suresh2022shapemap} can resolve such issue.}

\revised{Another direction is to incorporate priors or explicit inertial sensing from IMUs to improve pose tracking accuracy. Example priors include assuming approximately planar motion during contact, imposing a constant velocity prior, and incorporating object dynamics models. Incorporating these into the pose graph optimization could further improve the accuracy of GelSLAM.}

\revised{Finally, while GelSLAM supports flexible, in-the-wild human operation without requiring a robotic platform, manual scanning can lead to redundant exploration, as it is difficult to identify which regions remain untouched. A natural extension is autonomous tactile exploration, where a robot actively plans the next touching locations and revisits uncertain regions to improve scanning efficiency and accuracy. Extending the system to multi-fingered tactile sensing offers another avenue to accelerate coverage and further improve efficiency.}

%\mytodo{After including the constant velocity prior, revise this section to mention the existance of such prior.}

\section{Conclusions}

In this work, we introduce GelSLAM, a novel tactile-only system for long-horizon object tracking and high-fidelity object-level 3D reconstruction from tactile sequences. \revised{Our main contribution is a set of tactile-specific SLAM components, including NormalFlow failure detection, keyframe selection, and loop detection, together with their system-level integration into a well-engineered SLAM system.} By leveraging differential representations, GelSLAM robustly captures spatial relationships between tactile readings, even under challenging conditions such as large rotations, minimal overlap, and low surface texture. These capabilities support robust loop closure and tactile relocalization, enabling the system to correct drift and maintain global spatial consistency over time. \revised{Unlike prior systems limited to hundreds of frames and can frequently include false loop detections, GelSLAM scales to tens of thousands of frames with zero false loops.} As a result, GelSLAM overcomes the inherently local nature of touch and achieves long-horizon tracking as well as building coherent global reconstructions with detailed geometry. 

In experiments, we show that GelSLAM runs online and maintains accurate tracking across a wide range of everyday objects, effectively correcting drift over long horizons. We also demonstrate, both qualitatively and quantitatively, that GelSLAM reconstructs detailed object-level 3D geometry with high accuracy. Using GelBelt, we further show that our system can reconstruct large objects such as a tree trunk while preserving fine surface details. To the best of our knowledge, GelSLAM is the first tactile-only system to achieve robust and accurate object-level 3D reconstruction at this level of fidelity.

\revised{To conclude, GelSLAM’s real-time, long-horizon object tracking capability can enhance robot manipulation performance and potentially reshape how robots utilize touch. Beyond tracking, its high-fidelity reconstruction performance demonstrates that tactile sensing can serve as a practical reconstruction modality, particularly for applications requiring high precision under heavy occlusion or involving transparent objects. We believe GelSLAM opens a new direction in tactile perception and manipulation by extending touch from a local sensing modality to one capable of global, long-horizon spatial understanding.}

\section*{Acknowledgments}
We thank Daniel McGann and Montiel Abello for insightful discussions. We are grateful to the CMU AI Maker Space and Greg Armstrong for support with collecting the tracking dataset. We also thank Jingyi Xiang and Jui-Te Huang for helpful comments on the manuscript.

\bibliographystyle{plainnat}
\bibliography{references}

\end{document}